\pdfoutput=1

\documentclass[11pt]{article}

\usepackage[preprint]{acl}

\usepackage{times}
\usepackage{latexsym}

\usepackage[T1]{fontenc}

\usepackage[utf8]{inputenc}

\usepackage{microtype}

\usepackage{inconsolata}

\usepackage{multirow}
\usepackage{graphicx}
\usepackage{enumitem}
\usepackage{subfigure}
\usepackage{hyperref}
\usepackage{algorithm}
\usepackage{algorithmic}
\usepackage{amsfonts}
\usepackage{diagbox}
\usepackage{tcolorbox}
\title{LLM-Human Pipeline for Cultural Grounding of Conversations}

\author{Rajkumar Pujari \and Dan Goldwasser\\
  Purdue University, West Lafayette, USA\\
  \texttt{\{rpujari,dgoldwas\}@purdue.edu}\\}

\begin{document}
\maketitle
\begin{abstract}
Conversations often adhere to well-understood social norms that vary across cultures. For example, while \textit{addressing work superiors by their first name} is commonplace in the Western culture, it is rare in Asian cultures. Adherence or violation of such norms often dictates the tenor of conversations. Humans are able to navigate social situations requiring cultural awareness quite adeptly. However, it is a hard task for NLP models.

In this paper, we tackle this problem by introducing a \textit{Cultural Context Schema} for conversations. It comprises (1) conversational information such as emotions, dialogue acts, etc., and (2) cultural information such as social norms, violations, etc. We generate $\sim$110k social norm and violation descriptions for $\sim$23k conversations from Chinese culture using LLMs. We refine them using automated verification strategies which are evaluated against culturally aware human judgements. We organize these descriptions into meaningful structures we call \textit{Norm Concepts}, using an interactive human-in-the-loop framework. We ground the norm concepts and the descriptions in conversations using symbolic annotation. Finally, we use the obtained dataset for downstream tasks such as emotion, sentiment, and dialogue act detection. We show that it significantly improves the empirical performance.
\end{abstract}

\section{Introduction} \label{sec:intro}
%
%

  
Social norms define behavioral expectations shared across groups and societies~\cite{sherif1936psychology}. They can help explain the differences in the way people from different cultural backgrounds react to the same situation~\cite{triandis1994culture,finnemore1996norms}. \citet{hymes1972models} describe \textit{norms of interaction} as `shared rules that implicate the belief system of a community', capturing the importance of representing norms when, either human or AI-systems, attempt to make sense of social interactions from different cultures.  

Motivated by Large Language Models emergent reasoning abilities~\cite{wei2022emergent,srivastava2023imitation,camburu2018esnli,suzgun2022challenging} several recent works attempted to create repositories of cultural norms using novel prompting approaches followed by automated verification of the generated descriptions \citep{fung-etal-2023-normsage,li2023normdial,ziems-etal-2023-normbank}. However, these efforts had limitations such as using synthetic conversations or focusing on a handful of situations.  LLMs also tend to suffer from several reliability issues such as hallucinations \citep{rawte2023survey} or high sensitivity to prompt structure \citep{kojima2023-step-by-step,prystawski2023think,chain-of-thought-2023}.

These limitations motivate a more general and principled approach for \textit{Cultural Context Understanding}, which, we argue, should be viewed as a pragmatic reasoning task. Norms are situated in specific settings and are associated with the social roles participants play \citep{hare2003roles}. Different expectations can be associated with individuals based on attributes such as their status, profession, or gender. Furthermore, these expectations are situation-dependent, for example, changing when engaging in professional or leisure activities. 

To that end, in this paper, we propose a novel Cultural Context Grounding framework for conversations. We tackle three key problems. {First}, \textit{norm representation} capturing the multi-party situated social expectations. {Second}, norm induction, i.e., populating the suggested representation with norm concepts, emerging from conversational data, and associating them with relevant contextual information. {Finally}, grounding the norm concepts in conversational data at scale and creating a dataset of norm-schema aligned conversations.

Our norm representation solution is inspired by the notion of \textit{scripts}~\citep{schank2013scripts}, i.e., structured representations of expected activities for different roles relevant to a specific scenario, in our case mapping to social scenarios. Unlike past work, \citet{ziems-etal-2023-normbank}, that developed a schema representing expectations over situated \textit{actions}, our goal is to capture how social norms manifest in conversational behavior. To allow for pragmatic inferences mapping the norm definition to conversations, we intentionally separate between \textit{factual components}, capturing observed information about the settings and content of the conversation, and \textit{cultural norm components}, capturing the expected behaviors and impact of violating that expectation. Fig.~\ref{fig:schema_structure}(a) depicts this separation.

Our framework follows a multi-stage approach to obtaining cultural information for existing conversations and grounding this information in conversation-specific details. We focus mainly on relevant social norms and their violations within these conversations.
We then evaluate the usefulness of the created cultural context dataset both qualitatively and quantitatively. Fig. \ref{fig:schema_pipeline} presents an overview of our framework. We describe it in \S \ref{sec:schema}.

We leverage LLMs such as GPT-$3.5$ to generate a large but potentially noisy corpus of conversation-specific cultural information. Then, we leverage an \textit{interactive human-in-the-loop} process (Alg.~\ref{algo:concept_discovery}) that efficiently utilizes culturally proficient human annotation to organize this information into meaningful concepts. We further ground the generated descriptions in the conversation details, such as participants, their relationships, and their behavior using symbolic annotation. Following this, we experiment with automated verification strategies to filter the obtained cultural information at scale. Then, we organize the conversations and the obtained cultural information into a meaningful schema structure. Finally, we propose a neural graph schema model that leverages obtained data to improve the empirical performance on conversation understanding tasks such as emotion detection, sentiment detection, and dialogue act detection across multiple datasets. Overall, our contributions can be summarized as follows:
\begin{enumerate}[nolistsep]
    \item We propose a novel Cultural Grounding pipeline for conversation understanding using LLM \& culturally-aware human annotation.
    \item We introduce \textit{Norm Concepts} and employ a human-in-the-loop (HiL) framework to create human-validated concepts supported by data. 
    \item We leverage automated verification strategies to clean the LLM-generated data. We further symbolically ground the conversations in norm concepts. We create a high-quality, large-scale dataset for cultural understanding.
    \item We leverage the cultural schema information to improve downstream conversational task performance. We present meaningful visualizations and human evaluation experiments to showcase the quality.\footnote{Code and Data: \url{https://github.com/pujari-rajkumar/cultural-schema-naacl2025}}
\end{enumerate}


\section{Related Work} \label{sec:related}
\noindent \textbf{Social Norm Datasets:} A few recent works have leveraged LLMs such as GPT-3 to create useful cultural norm datasets using structure prompting approaches \citep{fung-etal-2023-normsage,li2023normdial,ziems-etal-2023-normbank}. While these works either focus on a small set or synthetically generated conversations, we generate $63k$ norm descriptions for $23.5k$ real conversations and ground them using a principled cultural grounding pipeline.

\noindent \textbf{Social Grounding:} \citet{pacheco-etal-2023-interactive} propose an interactive concept discovery for COVID-19 tweets. Several works address the tasks of principled grounding in social domain\citep{smith2018closing,roberts2019attempting,roy2021analysis,demszky2019analyzing,pujari2024we}. However, they mainly focus on social grounding in political settings. We focus on the cultural aspect of social context which is challenging to obtain explicit data.

\noindent \textbf{Automated Verification using LLMs:} Several previous works address automated annotation using LLMs and refining their generations.\citep{chiang2023can,li2021deus,bano2023exploring,bang2023multitask,yao2022react,hendrycks2020aligning}. We build upon the frameworks proposed by \citet{wu2023autogen} and \citet{arabzadeh2024assessing} to operationalize our annotation framework.

\section{Data} \label{sec:source_data}
\begin{table}[tbh]
\centering
\resizebox{200pt}{!}{
\begin{tabular}{ll}
\hline
\multicolumn{2}{l}{\textbf{Dataset: MPDD}}                                                                                                                                                                                                                                                                                \\ \hline
\multicolumn{1}{l|}{\textbf{Language:} Chinese}                                                                                                                                 & \textbf{\# Convs:} 4,141                                                                              \\
\multicolumn{1}{l|}{\multirow{2}{*}{\begin{tabular}[c]{@{}l@{}}\textbf{Description:} Conversations\\ from Chinese TV series scripts\end{tabular}}}                              & \textbf{\# Turns:} 25,546                                                                             \\
\multicolumn{1}{l|}{}                                                                                                                                                                            & \textbf{Tasks:} Emotion                                                                               \\ \hline
\multicolumn{2}{l}{\textbf{Dataset: CPED}}                                                                                                                                                                                                                                                                                \\ \hline
\multicolumn{1}{l|}{\textbf{Language:} Chinese}                                                                                                                                 & \textbf{\# Convs:} 11,832                                                                             \\
\multicolumn{1}{l|}{\multirow{2}{*}{\begin{tabular}[c]{@{}l@{}}\textbf{Description:} Textual dataset\\ with multi-modal features\\ from 40 Chinese TV shows\end{tabular}}}      & \textbf{\# Turns:} 132,723                                                                            \\
\multicolumn{1}{l|}{}                                                                                                                                                                            & \begin{tabular}[c]{@{}l@{}}\textbf{Tasks:} Dialogue Act,\\ Emotion, \& Sentiment \end{tabular}       \\ \hline
\multicolumn{2}{l}{\textbf{Dataset: LDC CCU ZH Text}}                                                                                                                                                                                                                                                                     \\ \hline
\multicolumn{1}{l|}{\textbf{Language:} Chinese}                                                                                                                                 & \textbf{\# Convs:} 6,763                                                                              \\
\multicolumn{1}{l|}{\multirow{2}{*}{\begin{tabular}[c]{@{}l@{}}\textbf{Description:} Chinese textual\\ dataset containing online forum \\ and chat conversations\end{tabular}}} & \textbf{\# Turns:} 98,821                                                                             \\
\multicolumn{1}{l|}{}                                                                                                                                                                            & \begin{tabular}[c]{@{}l@{}}\textbf{Tasks:} Norm Violation,\\ Emotion, \& Dialogue Act \end{tabular} \\ \hline
\end{tabular}
}
\caption{List of our raw data sources. ZH: Chinese; \#Convs: Conversations; \#Turns: Conversation Turns}
\label{tab:data_details}
\end{table}

\begin{figure*}[tbh]
    \centering
      \subfigure[Proposed Schema Structure]{\includegraphics[height=4.15cm]{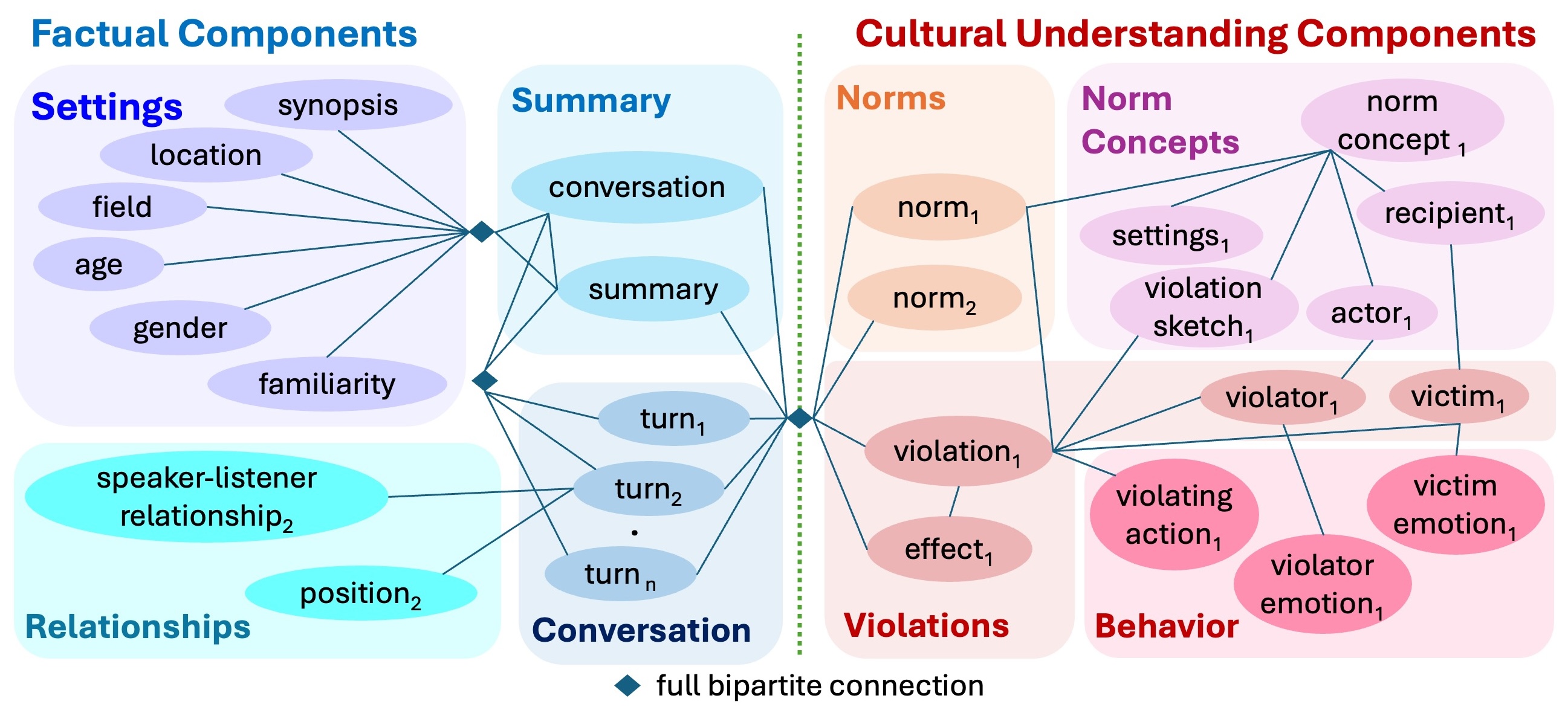}}\hfill
      \subfigure[\textit{Norm Concepts} \& \textit{Symbolic Explanations} example. $S_{*}$ denotes relevant pipeline stages from Fig. \ref{fig:schema_pipeline}.]{\includegraphics[height=4.15cm]{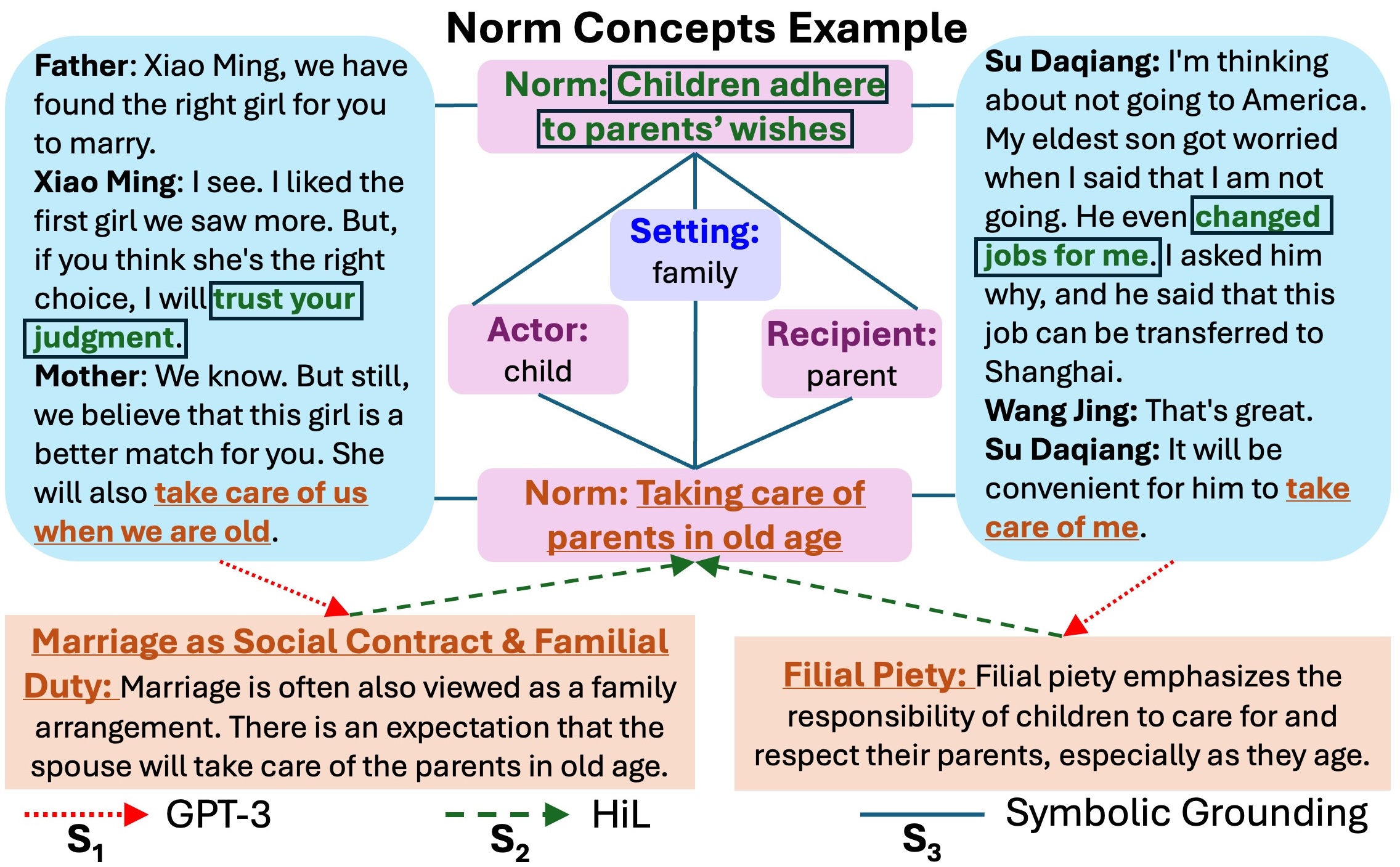}}\hfill
      \caption{Proposed \textit{Cultural Context Schema Structure for Conversations} with an example instance}
      \label{fig:schema_structure}
\end{figure*}

We build upon three existing dataset: MPDD \citep{chen-etal-2020-mpdd}, CPED \citep{chen2022cped}, and LDC CCU Chinese Text\footnote{\url{https://www.darpa.mil/program/computational-cultural-understanding}} datasets. These datasets consist of Chinese conversations annotated with emotions, sentiments, dialogue acts, and social norm violations. Each dataset is annotated for some of these tasks, but not all. Detailed statistics of the dataset are presented in Tab.\ref{tab:data_details}. We describe each dataset briefly below.

\noindent \textbf{MPDD Dataset:} \citep{chen-etal-2020-mpdd} This dataset contains 4,141 dialogues from Chinese TV series scripts. It is annotated with emotions, listeners, and speaker-listener relationships for each turn. The dataset contains 25,546 conversation turns. 

\noindent \textbf{CPED Dataset:} \citep{chen2022cped} CPED contains transcripts from 40 Chinese TV shows. It contains 11,832 conversations. It is annotated for emotions, sentiments, and dialogue acts. This dataset also provides multi-modal features such as face position, etc. We don't use them in this work. 

\noindent \textbf{LDC CCU ZH Text Dataset:} This dataset is a conversational understanding dataset from the DARPA CCU program. This dataset consists of text, audio, and video data in several languages. We use the text portion of the Chinese dataset for our experiments. It contains annotations for emotion, dialogue act, social norm violation status, and conversation change points. We focus on emotion and dialogue act identification tasks. This dataset also has metadata about settings, age of participants, familiarity, etc, annotated. The dialogue act annotation for this dataset is not complete. The annotation is focused only on a pre-determined set of dialogue acts. We deal with this by using \textit{other} class for unlabeled data.

As not all datasets have all components annotated, we train the Llama-3.1-70b-Instruct \citep{dubey2024llama} model to predict missing fields for each dataset using QLoRA \citep{dettmers2024qlora} fine-tuning. We predict relationships for CPED and LDC datasets. We predict LDC-style metadata for CPED and MPDD datasets.

\section{Cultural Context Grounding} \label{sec:schema}
Our high-level objective is to `\textit{conceptualize \& operationalize cultural understanding in conversational behavior}.' We identify three key steps to make progress towards this objective:
\setlist{nolistsep}
\begin{enumerate}[noitemsep]
    \item Formalize relevant cultural context for better conversational understanding.
    \item Obtain high-quality cultural context data at scale, leveraging native-culture expertise and ground the conversations in this context.
    \item Evaluate the obtained cultural context dataset for \textit{correctness} and \textit{usefulness}.
\end{enumerate}
First, we propose a graph-based schema structure for culturally enriched conversational representation. Our schema consists of two complementing segments: \textit{factual} components and \textit{cultural understanding} components. We present an overview of the schema structure in Fig.~\ref{fig:schema_structure} and a detailed description of the schema structure in \S\ref{sec:schema_structure}. 

\begin{figure*}[tbh]
    \centering
      \includegraphics[width=\textwidth]{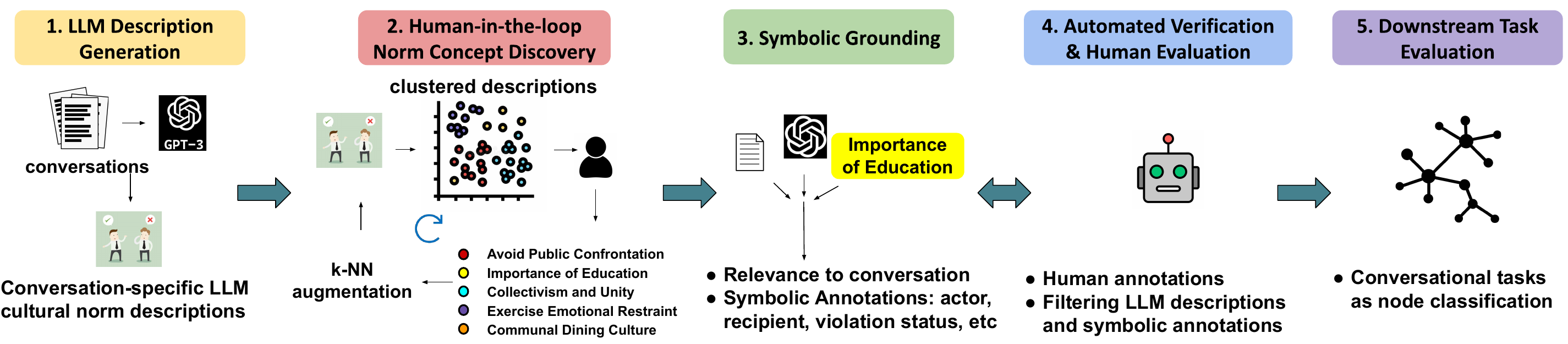}
      \caption{Proposed Cultural Context Grounding Pipeline for Conversations}
      \label{fig:schema_pipeline}
\end{figure*}

Then, we introduce a robust pipeline for obtaining cultural information for real conversations and grounding the conversations in it. We efficiently leverage (1) native-culture human expertise, (2) LLMs as knowledge elicitors, (3) LLMs as symbolic annotators, and (4) LLMs as multi-agent verifiers, in this pipeline. We obtain a large-scale corpus for $\sim$23k Chinese conversations from three existing datasets (Tab.~\ref{tab:data_details}). We present an overview of the proposed pipeline in Fig.~\ref{fig:schema_pipeline} and a detailed pipeline description in \S\ref{sec:grounding_pipeline}.

Finally, we evaluate (1) \textit{correctness} of the cultural information obtained using elicitor \& symbolic annotator LLMs, and (2) \textit{usefulness} of the schema for conversational understanding. We measure \textit{correctness} against human annotations (\S\ref{sec:framework_eval}). We evaluate the \textit{usefulness} of cultural schema data via conversational tasks such as emotion, sentiment, and dialogue act detection (\S\ref{sec:experiments}). We present full statistics of our pipeline data collection process in Tab.~\ref{tab:data_sources} in appendix \ref{sec:pipeline_data_statistics}.

\subsection{Schema Structure} \label{sec:schema_structure}

Context often dictates whether or not a specific behavior is considered normative. For example, while patronizing younger people might be commonplace in families in some cultures, it could be considered offensive in a professional setting. Especially if the younger person is a work superior. On the other hand, even within the same social setting, norms might vary circumstantially. While joking about an elder's forgetful nature might be okay in some situations, it could be considered insensitive if they are suffering from dementia.

This guides us toward two distinct genres of information that could influence conversational behavior: factual and cultural information. To account for this, we propose a schema structure with two distinct segments (Fig.~\ref{fig:schema_structure}(a)). The \textit{factual} segment of our schema consists of settings, summary, conversation, \& relationship components. \textit{Cultural Understanding} segment consists of social norms, violations, norm concepts, \& behavior components.

While factual information such as age group, location, and relationships is available for many datasets, descriptions of relevant social norms, their violations, and how they affect the conversation are harder to procure. We address this challenge by efficiently using LLMs and human annotation.

\subsection{Grounding Pipeline}\label{sec:grounding_pipeline}

Grounding in NLP usually refers to linking text or speech to real-world concepts such as entities, attitudes, etc \citep{chandu-etal-2021-grounding}. Conversational grounding can encompass various aspects, such as mutual beliefs, shared knowledge, assumptions, and so on \citep{clark1996using}. In this work, we focus specifically on the \emph{cultural knowledge} aspect of conversational understanding.

Obtaining exhaustive knowledge of a culture's social norms at scale is impractical. Therefore, we propose a bottom-up approach. We use real-world conversations to mine situation-dependent social norms, leveraging LLMs as \textit{knowledge elicitors}. Native-culture human annotators then create structured abstractions over these descriptions, which we refer to as \textit{norm concepts}. Building upon \citet{pacheco-etal-2023-interactive}, we devise an interactive framework that amplifies human judgments and scales them to the entire dataset. Then, we use LLMs as \textit{symbolic annotators} to ground conversations in human-generated norm concept structures.

As LLMs are susceptible to hallucinations and bias, we evaluate the obtained dataset and symbolic annotations against human judgments. Then, we further employ LLMs as multi-agent verifiers to significantly improve the quality of the dataset. We present an overview of the proposed pipeline approach in Fig.~\ref{fig:schema_pipeline}. We further discuss details of  each step of the pipeline in the following subsections.

\subsubsection{LLM Description Generation}\label{sec:llm_generation}
Two major challenges we face in building our cultural context dataset are (1) collecting structured knowledge about social norms as they manifest in real-world conversations and (2) scaling the descriptions of culturally nuanced behaviors observed in conversations. We address these challenges by leveraging existing conversational datasets and  LLMs as \textit{knowledge elicitors}, respectively. We provide raw conversations to the LLM and instruct it to explain observed behavior from a cultural perspective. For each conversation, we obtain descriptions of relevant social norms, potential violations, and their effects on the conversation trajectory and the participants. We present an example of norm descriptions in Fig.~\ref{fig:schema_structure}(b). We further provide the exact prompts and a full example of LLM outputs in the appendix \ref{sec:gpt_prompts}. We expect this step to be noisy as LLMs are susceptible to hallucinations. However, our primary focus is to obtain broad coverage of diverse cultural nuances that influence behavior in conversations. We evaluate the quality of LLM generations in \S\ref{sec:framework_eval}.


\subsubsection{HiL Norm Concept Discovery}\label{sec:norm_concept_discovery}
As noted in \citet{clark1996using}, humans actively draw from a \textit{`common ground'} when engaging in conversations. From a cultural standpoint, this common ground includes shared cultural beliefs. Humans often encounter social interactions where cultural awareness is practiced, which makes them adept at \textit{situating} conversations in \textit{cultural common ground}. Many aspects of this cultural common ground are highly general and serve us in a variety of situations. However, structured datasets that serve as a common ground for NLP models are hard to create solely using either human annotation or automated methods. Hence, in an attempt to create such a resource at scale, we employ culturally aware humans in an interactive human-in-the-loop (HiL) framework, which amplifies their judgments in a semi-automatic fashion.

\begin{algorithm}[tbh]
\resizebox{0.75\linewidth}{!}{ 
    \begin{minipage}{\linewidth}
        \caption{Interactive Norm Concept Discovery}
        \textbf{Input}: Conversations and their norm descriptions\\
        \textbf{Outputs}: Discovered \textit{norm concepts} with \textit{symbols}, many-to-one mapping: $\langle$desc, conv$\rangle$ $\rightarrow$ concept
        
        \begin{algorithmic}[1] 
            \setlength{\itemsep}{0pt} 
            \STATE Cluster norm descriptions using $k$-means.
            \STATE Cultural experts create \textit{norm concepts} by selecting 5-10 samples of closely related descriptions and providing symbolic structure.
            \STATE Perform $k$-NN augmentation of unmapped norm descriptions to each norm concept.
            \STATE Experts inspect augmented clusters and mark $5$-$10$ good \& bad examples for each concept.
            \STATE Re-assign norm descriptions to concepts using good \& bad cluster centers.
            \STATE Go back to step $1$ with the remaining unmapped norm descriptions.
        \end{algorithmic}
        \label{algo:concept_discovery}
    \end{minipage}
}
\end{algorithm}

Among LLM-generated cultural data, we observe overlapping norm descriptions across conversations, with minor variations. More interestingly, we also find closely related descriptions that can be grouped under the same theme. Consider the example in Fig.~\ref{fig:schema_structure}(b). Norm descriptions `\textit{marriage as a familial duty: spouse is expected to take care of parents in old age}' and `\textit{filial piety towards parents in old age}' are both defined by the common theme `\textit{care of parents in old age as child's responsibility}.' 

\begin{table}[htb]
    \centering
    \resizebox{180pt}{!}{
        \begin{tabular}{l}
            \hline \hline
             \textbf{Concept Name:} Respect For Authority\\\hline
             \textbf{Description:} Respecting hierarchies in family, \\
             professional, \& organizational settings. It involves\\
             individuals respecting the decisions, suggestions,\\
             orders, and advice from those in higher positions\\\hline
             \textbf{Settings:} workplace, family, organizations \\\hline
             \textbf{Violation Sketch:} Behavior that intentionally\\
             contradicts the expectations and decisions of the\\
             people in charge\\\hline
             \textbf{Actors:} sub-ordinates or people in an
             inferior\\ social position such as students, children, etc\\\hline
             \textbf{Recipients:} people in a position of power\\
             or authority over other people\\\hline
            \hline
        \end{tabular}
    }
    \caption{Cultural expert annotation of symbolic structure for discovered norm concept \textit{Respect for Authority}}
    \label{tab:concept_explanation_example}
\end{table}

To capture such themes in a principled manner, we introduce \textbf{\textit{norm concepts}}. They are abstractions over cultural beliefs which influence several related behaviors. We associate them with symbolic explanations, thus creating structured representations. 

A norm concept is characterized by an activation setting, a violation sketch, and actor and recipient roles. \textit{Actor} role describes people expected to adhere to the norm concept. \textit{Recipients} perceive/experience the consequences of adherence or violation. An example concept, \textit{`Respect for Authority'}, is presented in Tab.~\ref{tab:concept_explanation_example}. The symbolic structure of norm concepts is shown in Fig.\ref{fig:schema_structure}(b). The goal of HiL methodology is to support the discovery process via interaction with data.

\citet{pacheco-etal-2023-interactive} propose an interactive concept learning framework for tweets. We extend their framework for norm concept discovery. We outline our process in Alg. \ref{algo:concept_discovery} and describe it below.

Norm concepts are \textit{validated by humans} and \textit{supported by data}. We create initial unnamed clusters of LLM norm descriptions. Humans inspect these clusters and create norm concepts by selecting $5$-$10$ closely related examples of the concept. Humans also define a symbolic structure for the norm concept. Then, we perform $k$-NN augmentation of the concepts using untouched descriptions.

Then, humans inspect the newly augmented samples for each norm concept and mark $5$-$10$ good and bad examples for each concept cluster. We re-perform $k$-NN augmentation for the untouched descriptions. Humans also create new norm concepts when they deem appropriate. We perform further iterations to discover more norm concepts. As the process progresses, the unnamed clusters evolve and reveal new concepts. This iterative process helps us reliably amplify human mapping decisions to the entire dataset. We discovered $35$ norm concepts during this phase with a coverage of $64\%$ over $67k$ norm descriptions we collected. Annotation interface screenshots are presented in appendix \ref{sec:annotation_gui}. We evaluate the results of the concept discovery process using human evaluation in $\S$\ref{sec:framework_eval}. All annotators are CSS graduate students. We discuss this in detail in appendix \ref{sec:annotation_guidelines}.



\subsubsection{Symbolic Grounding} \label{sec:symbolic_annotation}
The HiL process creates norm concepts and maps them to conversation-specific descriptions. However, to fully ground the conversation in a norm concept, we must also align the conversation to the concept's symbolic structure. Hence, we leverage LLMs as \textit{symbolic annotators} to identify instantiations of concept symbols in the conversation.

We provide the symbolic annotator LLM with a conversation, LLM-generated descriptions, and associated norm concept structure. We first ask to verify the \textit{relevance} of the description to the conversation (filtering hallucinations) and the \textit{correctness} of description-concept mapping (filtering incorrect HiL mappings). Then, we ask it to annotate violation status, actor roles, recipient roles, and violation details, if applicable. We provide the exact prompts and outputs in appendix \ref{sec:symbolic_annotation_prompts}. We evaluate violation status annotations against human annotation in \S\ref{sec:framework_eval}.

\subsubsection{Automated Verification} \label{sec:automated_verification}


\begin{figure}[tbh]
    \centering
    \includegraphics[width=0.9\linewidth]{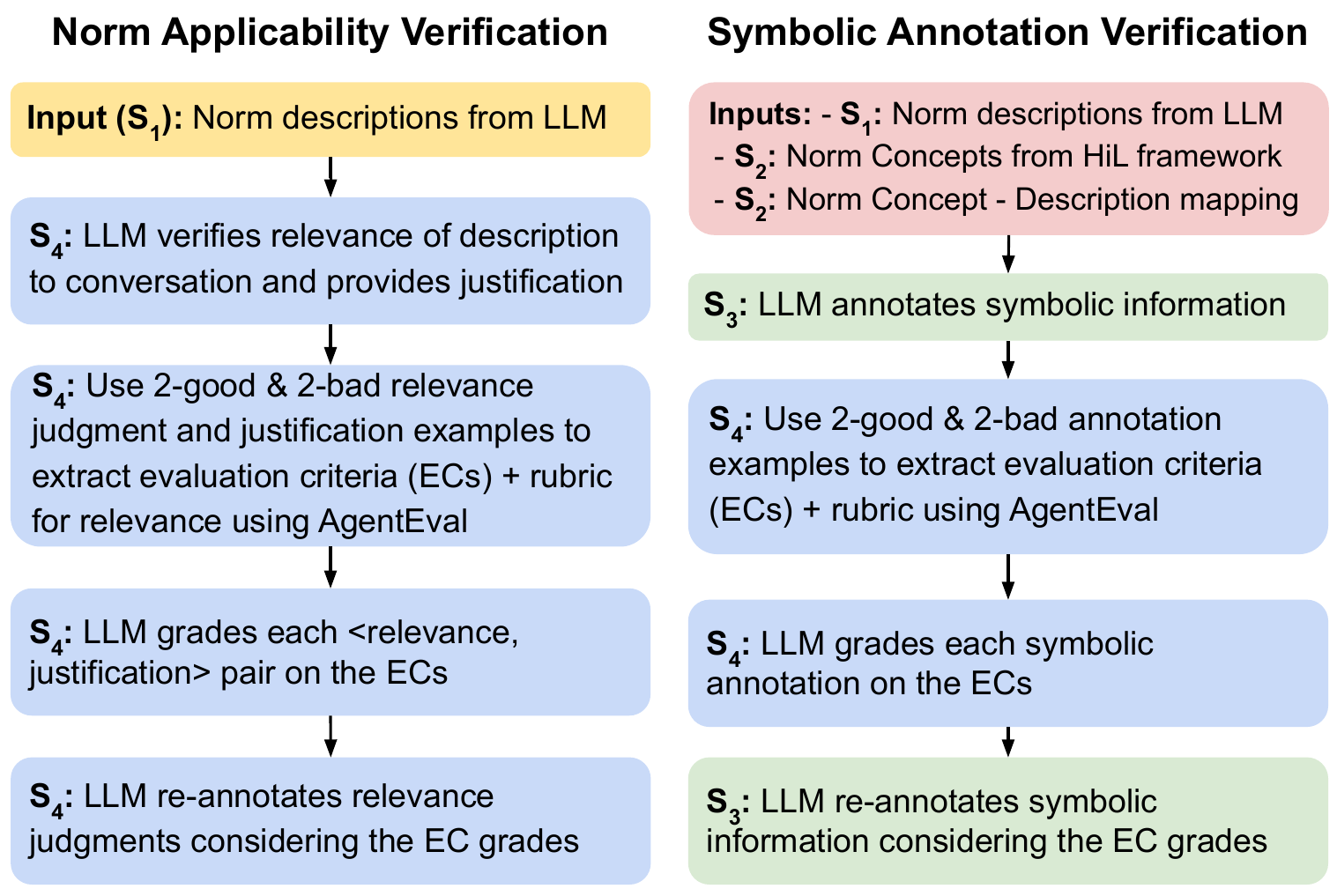}
    \caption{Automated Verification Flowcharts. $S_{*}$ denotes pipeline stages from Fig.~\ref{fig:schema_pipeline}}
    \label{fig:automated_verification}
\end{figure}

As we use LLMs and semi-automated decision amplification, the dataset is prone to noise from hallucinations and incorrect mappings. Hence, we devise techniques to refine the dataset in a systematic manner. \citet{fung-etal-2023-normsage} have shown that self-verification improves the LLMs output quality significantly. We aim to further improve this process by leveraging LLMs as multi-agent verifiers. To achieve this, we employ \textit{AgentEval} tool proposed by \citep{arabzadeh-clarke-2024-frechet}. We evaluate the initial, self-verified, and AgentEval-refined datasets against human judgments in \S\ref{sec:framework_eval}. We refine both the LLM descriptions and symbolic grounding annotations using AgentEval.

\noindent\textbf{Self-Verification} is the technique of prompting an LLM to re-consider a judgment. This is shown to be an effective technique in improving LLM response quality \citep{weng-etal-2023-large, fung-etal-2023-normsage}. We employ this technique to refine descriptions, HiL mappings, and symbolic grounding of conversations. Exact prompts used for self-verification are in appendix \ref{sec:gpt_prompts}.

\noindent\textbf{Multi-Agent Verification} We further improve our refinement process via a principled and highly interpretable multi-agent verification framework. AgentEval \citep{arabzadeh2024assessing} is a multi-agent framework for evaluating task utility. It takes the task description and 1-2 examples of successful and unsuccessful task runs as input. It generates a categorical rubric of criteria that helps determine the success/failure of future task runs.

The framework uses critic, quantifier, and verifier agents. The critic agent generates several criteria to evaluate the examples, the quantifier agent rates the examples on the criteria, and the verifier agent checks the robustness of each generated criterion. We add an evaluator agent on top of the criteria judgments to decide whether a data sample should be retained or not. We show the workflow charts in Fig.~\ref{fig:automated_verification}. We use instances of GPT-4o-mini for each agent in the framework. We provide the criteria generated for each evaluation workflow in appendix \ref{sec:agenteval_criteria}. 

\subsection{Comparison with Existing Norm Datasets}
Three of the prominent existing works that address social norms in conversation are NORMSAGE \citep{fung-etal-2023-normsage}, NORMDIAL \citep{li2023normdial} and NORMBANK \citep{ziems-etal-2023-normbank}. 

\noindent \textbf{NORMSAGE} presents a framework to obtain norms using LLMs and then perform self-verification. Their main contribution is the pipeline rather than a dataset. In contrast, we gather norm descriptions, operationalize a structured organization and filtering pipeline that efficiently uses and amplifies culturally proficient human annotation.

\noindent \textbf{NORMDIAL} is closest to our work as they propose a human-in-the-loop approach with LLM to obtain situation-specific norms. However, we introduce \textit{norm concepts} that aggregate situation-specific descriptions into more general behavioral concepts. We also present robust automated verification, which is evaluated using multi-point human evaluation. The scale of our dataset is larger.

\noindent \textbf{NORMBANK} uses situational attributes that align with the factual segment of our schema. However, unlike our work, these norms are not derived from real conversations but are instead defined by a list of high-level scenarios. Our dataset focuses on capturing nuanced human behavior within a specific culture, whereas NORMBANK aims for broader, more general cultural coverage.






\section{Qualitative Evaluation} \label{sec:framework_eval}

\begin{table}[tbh]
\resizebox{\columnwidth}{!}{
\begin{tabular}{lcc|cc|cc}
\hline
                                        & \multicolumn{2}{c|}{\textbf{Relevance}}            & \multicolumn{2}{c|}{\textbf{Mapping}}              & \multicolumn{2}{c}{\textbf{Violations}}            \\ \cline{2-7} 
                                        & \multicolumn{1}{c|}{\textbf{qual}} & \textbf{ret}  & \multicolumn{1}{c|}{\textbf{qual}} & \textbf{ret}  & \multicolumn{1}{c|}{\textbf{qual}} & \textbf{ret}  \\ \hline
\multicolumn{1}{l|}{\textbf{LLMs}}      & \multicolumn{1}{c|}{81}            & -             & \multicolumn{1}{c|}{91}            & -             & \multicolumn{1}{c|}{60.3}          & -             \\ \hline
\multicolumn{1}{l|}{\textbf{Self-Ver}}  & \multicolumn{1}{c|}{82.2}          & 73            & \multicolumn{1}{c|}{93.4}          & 85.6          & \multicolumn{1}{c|}{64.3}          & 74            \\ \hline
\multicolumn{1}{l|}{\textbf{MultiAgent}} & \multicolumn{1}{c|}{\textbf{88.4}} & \textbf{91.3} & \multicolumn{1}{c|}{\textbf{94.8}} & \textbf{93.8} & \multicolumn{1}{c|}{\textbf{66.1}} & \textbf{81.4} \\ \hline
\end{tabular}
}
\caption{Summary of human evaluation results of pipeline and refinement techniques. qual(ity) - \% of correct samples in the refined data; ret(ention) - \% of correct samples which passed refinement \label{tab:verification_results}}
\end{table}

\noindent Our cultural context dataset is prone to reliability issues from many sources, such as LLM hallucinations and scaling errors in the HiL process. Hence, we employ automated verification to rectify these errors. Now, we evaluate the quality of the obtained data and the refinement techniques via human evaluation. This allows us to quantify the reliability of the cultural context dataset as a resource. To that end, we obtain human judgments for several aspects of the dataset on a sampled subset. We observe that the automated verification-based filtering strategies significantly improve the dataset's quality. AgentEval's criteria-based refinement outperforms the self-verification strategy in all aspects.

Ideally, refinement techniques should retain \textit{good} samples while discarding \textit{bad} samples. Hence, we focus on two metrics for our evaluation: \textit{quality} and \textit{retention}. \textit{Quality}, which is analogous to precision, is defined as the percentage of \textit{good} samples in the post-refinement data. \textit{Retention}, analogous to recall, is defined as the percentage of original \textit{good} samples retained after refinement.

We randomly sample $726$ description across $239$ conversations for human evaluation annotation. Out of $726$ examples, $580$ are mapped to created norm concept structures. We focus our evaluation on three aspects: LLM hallucinations, symbolic annotation, and HiL scaling errors. Hence, we ask humans to (1) judge the \textit{relevance} of the LLM-generated description to the conversation, (2) judge whether the mapping of the norm concept to the description is accurate, and (3) judge whether or not the norm concept is violated in that particular conversation. Each of the three decisions is a binary yes/no answer. We use three native culture annotators for these tasks. All annotators are graduate students who are CSS researchers as well.

We evaluate the initial LLM generations from \S\ref{sec:llm_generation}, self-verification filtered, and AgentEval filtered datasets against human judgment data. We present the results in Tab.~\ref{tab:verification_results}. We observe that the quality of the dataset improves upon self-verification in all three aspects. It further improves significantly upon multi-agent verification. We reduce hallucinations in norm description generations from 19\% to 11.6\%, improve description-norm concept mapping quality from 91\% to 94.8\%, and improve violation status quality from 60.3\% to 66.1\% using multi-agent verification refinement. We also note that multi-agent verification retains a significantly higher portion of data than self-verification. For norm descriptions, we retain 91.3\% of correct descriptions (vs. 73\% by self-verification); for norm concept mapping, we retain 93.8\% (vs. 85.6\%); and for violation judgments, we retain 81.4\% (vs. 74\%) data.

\noindent \textbf{IAA:} We also measure inter-annotator agreement to quantify the subjectivity of these tasks. For concept mapping, we obtain Krippendroff's alpha of $0.61$, which points to moderate to strong agreement. For hallucinations, we obtain $0.74$, and for violation status, we obtain $0.68$. We also ask annotators to rate k-NN augmentations on how relevant they are to the assigned norm concept on a Likert scale of 1-5. This resulted in an average Likert score of $4.11$. These results demonstrate high agreement and a high success rate of k-NN augmentation. We further present interesting qualitative visualizations of norm concepts in appendix \ref{sec:norm_concept_visualization}.

\section{Downstream Task Evaluation} \label{sec:experiments}
\noindent Our experiments aim to evaluate the \textit{usefulness} of cultural information and graph-based schema. We use empirical performance on conversational understanding tasks as a proxy for \textit{usefulness}. We focus on $2$ classes of models: no-context models and cultural context models. Cultural context models are further divided into $2$ types: models that consume cultural context as \textit{text} and as \textit{a graph}. We use $3$ models for our experiments: RoBERTa \citep{cui2021pre}, LLama-3.1 \citep{dubey2024llama}, and our ConvGraph model. We design a graph model that uses PLM as a node encoder and leverages the schema structure in Fig.\ref{fig:schema_structure}(a). We briefly discuss the datasets and tasks and then models used.

\noindent \textbf{Datasets:} We perform experiments on $3$ tasks across $3$ existing datasets. We conduct experiments with MPDD, CPED, and LDC CCU Chinese datasets (\S\ref{sec:source_data}). We use emotion detection (all $3$ datasets), sentiment detection (CPED), and dialogue act identification (LDC CCU, CPED) tasks to benchmark the models.

\noindent \textbf{Tasks:} For emotion detection, MPDD is annotated using $7$ emotion labels, CPED with $13$, and LDC CCU ZH with $9$ labels. Dialogue act identification in CPED is a $19$ class task, and the LDC CCU Chinese dataset uses $10$ labels. CPED sentiment is a $3$-way annotation. We use the original train/validation/test splits for CPED and MPDD datasets. For the LDC CCU Chinese dataset, we use the LDC2022E18 release as the train set, the LDC2023E01 release as the validation set, and the LDC2023E20 release as the test set. In this dataset, \textit{neutral} label for emotion detection and \textit{other} for dialogue act identification are over-represented. We also find that several samples from these classes are missed annotations. Hence, to avoid skewing the models, we down-sample these classes in all data splits to $1$\% of the actual labels. This makes these classes roughly the same size as the other classes.

\subsection{Models} \label{sec:downstream_models}

We experiment with 3 models: RoBERTa, LLama-3.1-8B-Instruct, \& our ConvGraph model.

\noindent \textbf{RoBERTa:} We fine-tune the RoBERTa-Chinese-WWM-base \citep{cui2021pre} model on each task as sequence classification. We provide all the previous turns in the conversation and the current turn as inputs and predict class labels for respective tasks. We use cross-entropy loss to train the model. We use loss weighting to deal with class imbalance. Further details are provided in appendix \ref{sec:reproducibility}. We use the HuggingFace transformers library \citep{wolf-etal-2020-transformers} for all our experiments.

\noindent \textbf{LlaMa-3.1-8B-Instruct}: We perform QLoRA \citep{dettmers2024qlora} fine-tuning of state-of-the-art LLM, Llama-3.1-8B-Instruct model using the same inputs as RoBERTa-Chinese-WWM model for the no-context experiments. Then, for cultural context experiments, we augment both factual component information and cultural information as text. We provide the exact input format in appendix \ref{sec:downstream_setup}. 

\noindent \textbf{ConvGraph Model:} For the no-context model, we create a graph from only conversational turns. The graph consists of one node with the conversation text. This node is connected to several child nodes with one turn each. We perform tasks as node classification. We use RoBERTa-Chinese-WWM-base to encode the dialogue and turn nodes. We use DGL \citep{wang2019dgl} library to implement the graph model. We use two GraphSAGE \citep{hamilton2018inductive} layers on top of the PLM encoders and then pass the node embedding to a final GraphSAGE layer for final task classification.

For the ConvGraph + cultural context model, we use the graph structure presented in Fig.~\ref{fig:schema_structure}(a). To encode cultural context nodes, which are in English, we use RoBERTa-base \citep{liu2019roberta} as node encoder. We represent all edges as bi-directional edges. In this model, the contextual nodes such as norm concepts, relationships, settings, etc, are shared across conversations. This makes the entire data split (train/valid/test) into a single graph. We use randomly sample a neighborhood of $10$ for each node during training and inference to make the computation tractable. 



\section{Results} \label{sec:results}

\begin{table}[!tbhp]
\resizebox{\columnwidth}{!}{
\begin{tabular}{lllllll}
\hline
\multicolumn{1}{c|}{\multirow{3}{*}{\textbf{\begin{tabular}[c]{@{}c@{}}Task/\\ Model\end{tabular}}}} & \multicolumn{1}{c|}{\textbf{MPDD}}  & \multicolumn{3}{c|}{\textbf{CPED}}                                                                              & \multicolumn{2}{c}{\textbf{LDC CCU}}                                    \\ \cline{2-7} 
\multicolumn{1}{c|}{}                                                                                & \multicolumn{1}{c|}{\textbf{Em}}    & \multicolumn{1}{c|}{\textbf{Em}}    & \multicolumn{1}{c|}{\textbf{Sent}}  & \multicolumn{1}{c|}{\textbf{DA}}    & \multicolumn{1}{c|}{\textbf{Em}}    & \multicolumn{1}{c}{\textbf{DA}}   \\ \cline{2-7} 
\multicolumn{1}{c|}{}                                                                                & \multicolumn{1}{c|}{\textbf{w-F1}}  & \multicolumn{1}{c|}{\textbf{w-F1}}  & \multicolumn{1}{c|}{\textbf{w-F1}}  & \multicolumn{1}{c|}{\textbf{w-F1}}  & \multicolumn{1}{c|}{\textbf{w-F1}}  & \multicolumn{1}{c}{\textbf{w-F1}} \\ \hline
\multicolumn{7}{l}{\textbf{No Context Models}}                                                                                                                                                                                                                                                                                         \\ \hline
\multicolumn{1}{l|}{RoBERTa}                                                                         & \multicolumn{1}{l|}{61.13}          & \multicolumn{1}{l|}{20.89}          & \multicolumn{1}{l|}{46.69}          & \multicolumn{1}{l|}{55.40}          & \multicolumn{1}{l|}{56.50}          & 64.84                             \\ \hline
\multicolumn{1}{l|}{ConvGraph}                                                                           & \multicolumn{1}{l|}{61.81}          & \multicolumn{1}{l|}{21.42}          & \multicolumn{1}{l|}{47.66}          & \multicolumn{1}{l|}{56.28}          & \multicolumn{1}{l|}{54.83}          & 65.62                             \\ \hline
\multicolumn{1}{l|}{\begin{tabular}[c]{@{}l@{}}Llama-3.1-\\ 8B-Instruct\end{tabular}}                & \multicolumn{1}{l|}{45.48}          & \multicolumn{1}{l|}{24.73}          & \multicolumn{1}{l|}{53.91}          & \multicolumn{1}{l|}{54.15}          & \multicolumn{1}{l|}{60.11}          & 67.15                             \\ \hline
\multicolumn{7}{l}{\textbf{Cultural Context Models}}                                                                                                                                                                                                                                                                                   \\ \hline
\multicolumn{1}{l|}{\begin{tabular}[c]{@{}l@{}}Llama-3.1-\\ 8B-Instruct\\ + Context\end{tabular}}    & \multicolumn{1}{l|}{48.40}          & \multicolumn{1}{l|}{\textbf{29.81}} & \multicolumn{1}{l|}{\textbf{55.00}} & \multicolumn{1}{l|}{\textbf{60.16}} & \multicolumn{1}{l|}{\textbf{62.16}} & 68.28                             \\ \hline
\multicolumn{1}{l|}{\begin{tabular}[c]{@{}l@{}}ConvGraph\\ + Context\end{tabular}}                       & \multicolumn{1}{l|}{\textbf{64.34}} & \multicolumn{1}{l|}{21.65}          & \multicolumn{1}{l|}{49.90}          & \multicolumn{1}{l|}{57.24}          & \multicolumn{1}{l|}{57.97}          & \textbf{71.74}                    \\ \hline
\end{tabular}
}
\caption{Results on test sets\label{tab:results}. We report weighted F1 scores. Em-emotion, Sent-sentiment, DA-Dialogue Act.}
\end{table}

\noindent We present the results on the test sets of each task of our experiments in Tab.\ref{tab:results}. We observe that the cultural context models outperform no-context models significantly in all the tasks. We report the frequency weighted-F1 score metric. 

We mainly note the performance of cultural context models. Both Llama-3.1-8B and ConvGraph models perform significantly better when the cultural context information is augmented. It is interesting to note that despite significant pre-training and post-training on extensive data, llama-3.1 models still benefit significantly from cultural context information. For the llama-3.1-8B model, performance on CPED emotion improves from 24.73 to 29.81, and CPED dialogue act performance improves from 54.15 to 60.16 F1. 

It is also interesting to observe that the ConvGraph + cultural context model performs best on two tasks: MPDD emotions and LDC CCU dialogue acts. ConvGraph model only has $\approx$ 500M parameters as opposed to Llama-3.1-8B.

It is intriguing to note that the Llama-3.1-8B model significantly lags behind even the RoBERTa baseline on the MPDD emotion task. We posited that this might be due to a lack of robust multi-lingual capabilities. Hence, we trained the Llama-3.1-70b model on this task as well. But, it also reaches only 53.41 F1 on the task. We conclude that this might be due to the nature of the Llama class of models pre-training, post-training, and the annotation distribution for this particular dataset.


\section{Conclusion} \label{sec:conclusion}
We propose a novel cultural context schema grounding pipeline. We introduce \textit{Norm Concepts} which abstract over cultural beliefs. Using the pipeline and LLMs, we create high-quality cultural context data and perform human evaluation. Finally, we show that the dataset improves conversational understanding empirically.


\section*{Acknowledgments} \label{sec:acknowledgment}
We thank the reviewers for their insightful comments that helped improve the paper. This work was supported by NSF CAREER award IIS-2048001 and the DARPA CCU program. The contents are those of the author(s) and do not necessarily represent the official views of, nor an endorsement by, DARPA, or the US Government.

\section*{Limitations} \label{sec:limitations}
We present a novel framework built upon LLMs and a human-in-the-loop approach. Both these approaches are prone to bias due to data and human components. Our evaluation is qualitative and hence relies on heuristic metrics. Our method requires expensive data collection and annotation protocol. We perform annotations for only one language. Cultural norm discovery using LLMs is limited due to the lack of depth on knowledge on some cultures for LLMs. Western bias due to training data might propagate into the dataset. This is a pioneering attempt to build large-scale datasets and hence requires careful usage of the data and technology. Our verification strategy also constitutes LLMs and hence that data still contains close to ~10\% noisy data event after validation.

LLM-generated cultural norms could potentially perpetuate stereotypes, contain hallucinations, bias, or other harmful distributional characteristics that could be challenging to detect and filter. In this work, we address these issues such as hallucinations, and stereotypes by (1) involving culturally proficient humans in the norm concept creation stage and (2) also performing principled automated filtering which is again evaluated against human annotation. In our pipeline, humans are not just aggregating the descriptions that they see, but rather they are using their cultural expertise to identify norm concepts that are prominent in their culture and are also supported by data. Despite these steps, the output is not necessarily completely absolved of these issues. On the other hand, human-generated datasets have also been shown to contain harmful biases \citep{hovy2021five,geva2019we,gautam2024blind,davidson2019racial,das2024investigating,doughman2022gender}. LLMs are powerful NLP tools that allow us to explore tasks that were not practical before their advent. A case in point is our paper. It would have been highly impractical or hugely expensive to collect culturally proficient human annotations for cultural norms at this scale without LLMs. We believe that a pragmatic approach to this issue would be to carefully document and evaluate biases at each stage of the pipeline and invest community effort in building de-biasing techniques, bias-free training approaches, and so on. This could potentially lead us to creating fair(er) systems. We believe that a pragmatic approach of learning to build better guard rails, mitigating the harmful effects, and being aware of these biases could be a more fruitful path forward.

\section*{Ethics Statement} \label{sec:ethics}
Cultural norm discovery using LLMs is limited due to the lack of depth on knowledge on some cultures for LLMs. Western bias due to training data might propagate into the dataset. This is a pioneering attempt to build large-scale datasets and hence requires careful usage of the data and technology.
Human annotation using cultural experts is a high-variance process as the experience of culture varies from region to region and person to person. We try to capture a high-level idea of culture in our work. It is possible that the data propagates biases in the society and hence requires careful usage. Our work is aimed to be a research artifact to help foster better models for cross-cultural interaction. This is by no means an end product to be used in large-scale applications.


\appendix
\raggedbottom
\newpage

\section{Norm Concept Visualization} \label{sec:norm_concept_visualization}
\begin{figure}[htb]
    \centering
    \includegraphics[width=0.9\columnwidth]{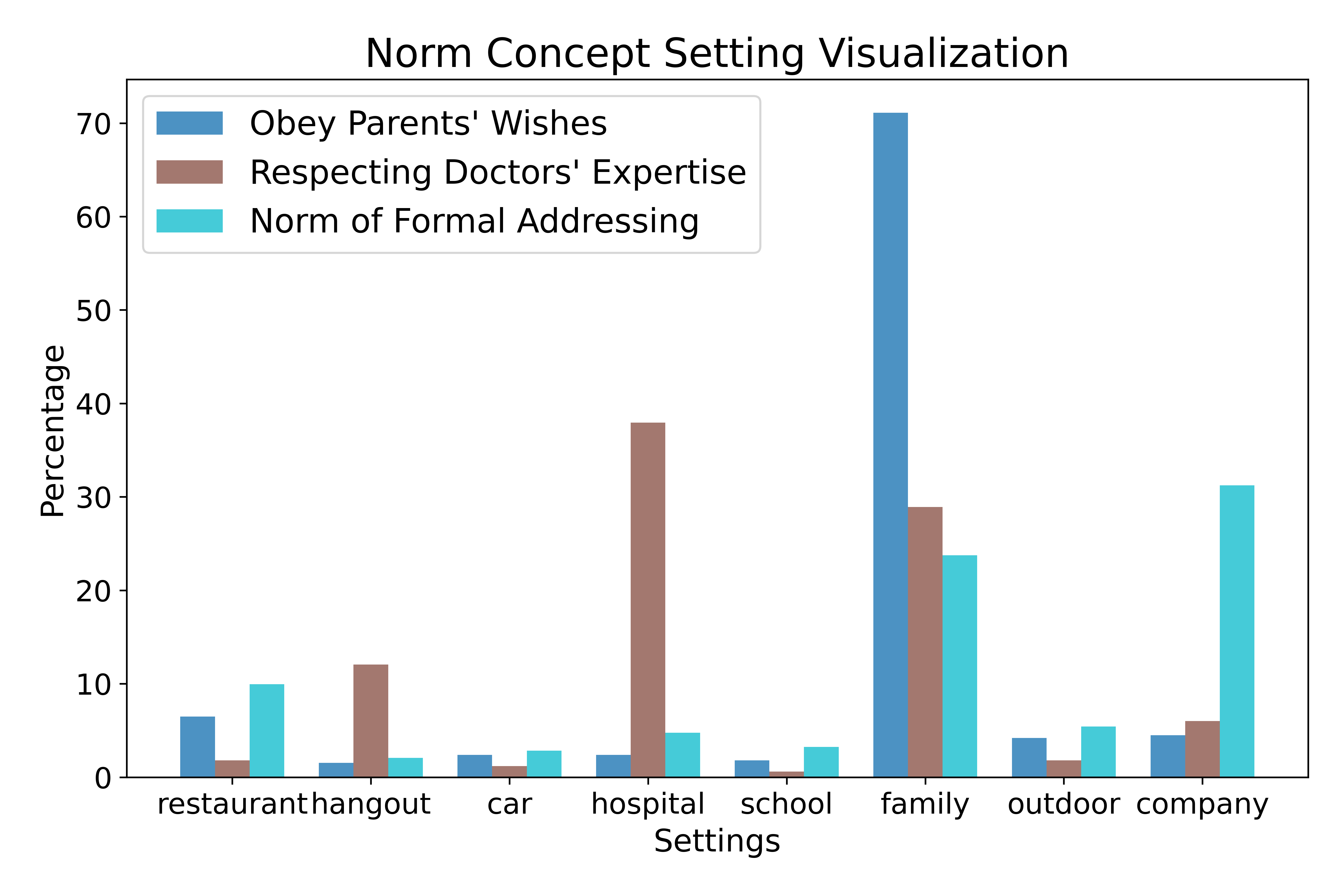}
    \caption{Comparison of Conversation Field Distribution Across Various Norm Concepts}
    \label{fig:field_visualiation}
\end{figure}

Further, we present a qualitative visualization of norm concept distribution over conversational settings in Fig.\ref{fig:field_visualiation}. We present the visualizations for 1) Norm of Formal Address, 2) Norm of Children obeying Parents' wishes and 3) Respecting the Doctor's expertise. We observe that norms are prominent in different fields. Formal address norm is highly prominent in a company setting, while children obeying parents' wishes is more prevalent in family settings. This is in accordance with the general expectations. This shows that the norm concepts capture interesting aspects of conversational context. Norm (3) is highly prominent in family and hospital settings. 

It is interesting to note that we only have field values at a conversational level, and hence, there could be multiple fields with the same conversation. The conversations are not segmented neatly on a scene-to-scene basis, especially in the LDC CCU dataset. This explains the noise distribution of certain norms. It is also noteworthy that the norm descriptions are generated based on conversational content. The field of the conversation alone doesn't always determine the conversational content. 

\section{Reproducibility} \label{sec:reproducibility}
For our downstream task experiments, we use a community cluster with several 80G and 40G A100 GPUs. We use a learning rate of 1e-4 for llama QLoRa training. We use r=16 with 8-bit quantization. For the graph model, we use the DGL library with distributed training using several A100 GPUs. Our llama training took 12 hours on average for each task. We ran several development iterations before arriving at the final prompt structure. The graph model training took 4 hours on average for each task, when augmented with cultural schema information. RoBERTa models take 10-30 minutes to converge in training depending on the size of the dataset. We will release all our code and datasets under MIT license upon acceptance. We extensively use ChatGPT, Claude, Github co-pilot for our coding requirements. We paid $\sim$400 USD for using OpenAI API for data collection and experiments. We use Grammarly to aid in draft improvement.

\section{Annotation Guidelines} \label{sec:annotation_guidelines}
All annotators used for both HiL concept discovery and human evaluation are graduate students, 3 of whom are also CSS researchers. We have a quid-pro-quo data annotation system in place in our lab. We contribute hours of annotations to other projects and we can get similar hours back when we need them. We use 3 native Chinese annotators, who were born and raised in mainland China for our annotations. We also use 2 non-Chinese annotators. They help in book-keeping tasks in the norm concept generation. They moderate and organize the interactive process sessions which are collaborative. The HiL process described in \S\ref{sec:norm_concept_discovery} is a collaborative annotation process. All the annotators discuss and reach a consensus before adding a norm concept to the database. In contrast, human evaluation annotation described in \S\ref{sec:framework_eval} is an individual annotation task. Hence, we report inter-annotator agreement for these judgments. This annotation is performed by native culture annotators.

\section{LLM Prompts and Generations}\label{sec:gpt_prompts}

\begin{tcolorbox}[colback=gray!10, colframe=gray!50, width=\columnwidth, boxrule=0.5mm, arc=2mm, auto outer arc, fontupper=\scriptsize]
\texttt{\textcolor{blue}{\textbf{Prompts used to extract Conversation-Specific Social Norms and Violations from GPT-3.5}}
\begin{enumerate}[nolistsep]
    \item Translate this conversation into English.
    \item List the people mentioned in the conversation and the social relationships between them.
    \item List the Chinese cultural norms applicable to this situation. Are there any cultural norm violations observed in this situation? If yes, list them. List the observed and potential effects by index for each violation.
    \item Summarize the conversation in 3-4 sentences.
\end{enumerate}}
\end{tcolorbox}

\begin{tcolorbox}[colback=gray!10, colframe=gray!50, width=\columnwidth, boxrule=0.5mm, arc=2mm, auto outer arc, fontupper=\scriptsize]
\textcolor{blue}{\texttt{\textbf{Input}}}\\
\texttt{\textbf{Translated Conversation:}}\\
\texttt{\textbf{Mrs. Zuo}: What is that foolish girl worth giving anything to? Zho Zpeng is such a person!}\\
\texttt{\textbf{Mr. Zuo}: Oh, wife, why do you always say things that are not conducive to unity? It's reasonable for him to go and give something to his classmate!}\\
\texttt{\textbf{Zho Zpeng}: Dad, Mom, I'm back!}\\
\texttt{\textbf{Mrs. Zuo}: I'm afraid you've been drugged and your soul has been seduced by that foolish classmate of yours!}\\
\texttt{\textbf{Mr. Zuo}: Wife, do you sound like a mother when you speak like that? Why do you always oppose their relationship?}\\
\texttt{\textbf{Zho Zpeng}: Dad, Mom, please stop talking. I have my own opinions on this matter. I know what I should do and what I shouldn't do. I will never act recklessly. Please trust me!}\\

\textcolor{blue}{\texttt{\textbf{LLM Description Generations}}}\\
\texttt{\textbf{Summary:}}\\
\texttt{The conversation revolves around Zho Zpeng, his parents, and his relationship with a classmate. Mrs. Zuo disapproves of the classmate and questions the worth of giving her anything, while Mr. Zuo defends his son's actions. Zho Zpeng expresses his independence and asserts that he knows what he should and shouldn't do, asking his parents to trust him.}\\

\texttt{\textbf{Norms:}}\\
\texttt{\textbf{Respect for parents:} Filial piety and respect for parents are highly valued in Chinese culture. Children are expected to listen to and obey their parents' opinions and decisions.}\\
\texttt{\textbf{Unity within the family:} Maintaining harmony and unity within the family is important in Chinese culture. Conflicts or disagreements that disrupt family unity are generally discouraged.}\\
\texttt{\textbf{Social relationships and obligations:} Chinese culture places importance on maintaining positive relationships and fulfilling social obligations, such as giving gifts to classmates or friends.}\\

\texttt{\textbf{Violations:}}\\
\texttt{\textbf{Disrespectful language:} Mrs. Zuo uses derogatory language to refer to Zho Zpeng's classmate, which shows a lack of respect for others.}\\
\texttt{\textbf{Opposition towards Zho Zpeng's relationship:} Mrs. Zuo continuously opposes the relationship between Zho Zpeng and his classmate, going against the cultural norm of respecting children's autonomy and choices in relationships.}\\

\texttt{\textbf{Effects:}}\\
\texttt{\textbf{Disrespectful language - Observed effect:} It can create tension and animosity between Mrs. Zuo and Zho Zpeng.}\\
\texttt{\textbf{Opposition towards Zho Zpeng's relationship - Observed effect:} It causes disagreement and arguments between Mrs. Zuo and Mr. Zuo, highlighting a lack of unity within the family.}
\end{tcolorbox}

\section{Symbolic Annotation Analysis} \label{sec:symbolic_annotation_prompts}

\begin{tcolorbox}[colback=gray!10, colframe=gray!50, width=\columnwidth, boxrule=0.5mm, arc=2mm, auto outer arc, fontupper=\scriptsize]

\textcolor{blue}{\texttt{\textbf{Symbolic Annotation \& Verification Prompts}}}

\texttt{As a two-step cultural and social norms analysis assistant, your task is to evaluate a provided conversation from the context of Chinese culture based on a given social norm and a corresponding norm concept. Your analysis should be comprehensive, considering factors such as age, relationships, settings (e.g., work, family, friends), and the topic of the conversation.\\}

\texttt{Steps for Analysis:\\}

\texttt{1. **Evaluate Social Norm and Norm Concept Compatibility**  
   - **Task:** Judge whether the provided social norm aligns with the given norm concept.
   - **Action:** State if the social norm **matches** or **doesn't match** the norm concept.
   - **Justification:** Provide a concise reason for your judgment.\\}

\texttt{2. **Assess Relevance to the Conversation**  
   - **Task:** ..\\}
  
\texttt{3. **Determine Social Norm Violation**  
   - **Task:** Judge whether the social norm was **adhered to** or **violated** in the conversation.
   - **Justification:** Provide a concise reason for your judgment.\\}
  
\texttt{4. **Annotate Conversation-Specific Details** \\}
   
   \texttt{- **Enactor Role:** ...\\
   - **Acceptor Role:** ...\\}

\texttt{5. **Violation Analysis** *(Only if a violation occurred)*  
   If the norm was violated, provide the following additional details:
   - **Violating Action:** A brief description of the action that caused the violation (e.g., "badmouthing parents").\\}
   
   \texttt{- **Violator Role:** ..\\}
   \texttt{- **Victim Role:** ..\\}
   \texttt{- **Violator Emotion:** ..\\}
   \texttt{- **Victim Emotion:** ..\\}

\texttt{Response Format:
Your response must adhere to the format below:\\}

\texttt{Social Norm - Norm Concept Compatibility: <match/doesn't match>  
Compatibility Justification: <short justification>\\}

\texttt{{Only if compatible}  
Relevance: <relevant/irrelevant>  
Relevance Justification: <short justification>\\}

\texttt{Enactor Role: <sitatuion specific social role, strictly not the name, of the person>  
Acceptor Role: <sitatuion specific social role, strictly not the name, of the person>\\}

\texttt{Violation Status: <adhere/violate>  
Violation Status Justification: <short justification>\\}

\texttt{{Only if a violation occurs}\\  
Violating Action: <short phrase> \\
Violator Role: <sitatuion specific social role, strictly not the name, of the person>\\  
Victim Role: <sitatuion specific social role, strictly not the name, of the person> \\ 
Violator Emotion: <one of 9 basic emotions>\\
Victim Emotion: <one of 9 basic emotions>}
\end{tcolorbox}

\begin{tcolorbox}[colback=gray!10, colframe=gray!50, width=\columnwidth, boxrule=0.5mm, arc=2mm, auto outer arc, fontupper=\scriptsize]
\textbf{Input:}\\

\texttt{\textbf{Conversation from Chinese culture:}}\\
\texttt{\textbf{Zuo Zhengpeng}: Lihua, how much longer until the baby is due?}\\
\texttt{\textbf{Xu Lihua}: Probably next week!}\\
\texttt{\textbf{Zuo Zhengpeng}: Oh, the baby is moving!}\\
\texttt{\textbf{Xu Lihua}: Look at you, how embarrassing!}\\
\texttt{\textbf{Xu Lihua}: What’s the rush? We still have a few days. We’ll talk about it then. You’re eager to become a father, aren’t you?}\\
\texttt{\textbf{Zuo Zhengpeng}: No! By then, it might be too late. It’s better to go to the hospital a few days early. Transportation here is inconvenient. If anything happens, we could be in big trouble.}\\

\texttt{\textbf{Social Norm:}}\\
\texttt{Respect for elders: It is common in Chinese culture to show respect to older family members, such as parents and grandparents.}\\

\texttt{\textbf{Norm Concept Name:} Respect for family elders}\\
\texttt{\textbf{Norm Concept Description:} Respecting the wisdom, experience, and authority of elder members in the family hierarchy.}\\
\texttt{\textbf{Norm Concept Potential Violation Sketch:} Showing disrespect and ignoring suggestions and advice of any elder members of the family.}\\
\texttt{\textbf{Norm Concept Scenario:} family}\\
\texttt{\textbf{Enactor Role:} any younger family member}\\
\texttt{\textbf{Acceptor Role:} elder family members such as parents, uncle, grandparents, etc.}
\end{tcolorbox}

\begin{tcolorbox}[colback=gray!10, colframe=gray!50, width=\columnwidth, boxrule=0.5mm, arc=2mm, auto outer arc, fontupper=\scriptsize]
\textbf{GPT Output:}\\

\texttt{\textbf{Norm-Concept Compatibility:} match}\\

\textcolor{red}{\texttt{\textbf{Relevance:} relevant}\\
\texttt{\textbf{Relevance Justification:} The conversation takes place within a family context, discussing the impending birth of a child, which inherently involves respect for family roles and responsibilities.}}\\

\texttt{\textbf{Enactor Role:} younger family member}\\
\texttt{\textbf{Acceptor Role:} elder family member}\\

\texttt{\textbf{Quality Judgment:} accurate}\\
\texttt{\textbf{Justification:} The annotations correctly identify the compatibility of the social norm with the norm concept, the relevance to the family context, and the adherence to the norm in the conversation, \textcolor{red}{reflecting a respectful interaction between the younger and elder family members}.}
\end{tcolorbox}

\section{Downstream Task Setup} \label{sec:downstream_setup}

\begin{tcolorbox}[colback=gray!10, colframe=gray!50, width=\columnwidth, boxrule=0.5mm, arc=2mm, auto outer arc, fontupper=\scriptsize]

\# Instruction and final <task-name> prompt are not provided for the RoBERTa model.\\

\textcolor{blue}{\textbf{\texttt{LlaMa-3.1-8B-Instruct, RoBERTa Models Input:}}}\\
 
\texttt{Instruction: You are a helpful assistant who predicts the <task-name> of the conversational turn in a Chinese conversation. You are given the cultural context surrounding the conversation, the prior conversation, and the current turn. Predict the <task-name> of the current turn. Choose one of <label set>.}\\

\texttt{Prior Conversation:\\
<turn 1>\\
<turn 2>\\}

\texttt{Current Turn:\\  
<turn-n>\\}

\texttt{**<task-name>**:}
\end{tcolorbox}

\begin{tcolorbox}[colback=gray!10, colframe=gray!50, width=\columnwidth, boxrule=0.5mm, arc=2mm, auto outer arc, fontupper=\scriptsize]

\textcolor{blue}{\textbf{\texttt{LlaMa-3.1-8B-Instruct + Cultural Context Model Input:}}}

\texttt{Instruction: You are a helpful assistant who predicts the <task-name> of the conversational turn in a Chinese conversation. You are given the cultural context surrounding the conversation, the prior conversation, and the current turn. Predict the <task-name> of the current turn. Choose one of <label set>.}\\

\texttt{Prior Conversation:\\
<turn 1>\\
<turn 2>\\}

\texttt{Conversation Settings:\\ 
Synopsis: ..\\
Speaker Count: ..\\  
Speaker Sex: ..\\
...\\}

\texttt{Relevant Cultural Information:\\  
1) **Norm Concept**:\\  
Theme: <theme-name>\\
Description: ...\\  
Settings: ..\\  
Violation Sketch: ...\\}

\texttt{Specific Norm:\\  
<norm-description>\\}

\texttt{2) **Norm Concept**:\\  
...\\}

\texttt{**Potential Violations**:\\  
1) <violation-description-1>\\
..\\}

\texttt{Current Turn:\\  
<turn-n>\\
**<task-name>**:} 

\end{tcolorbox}

\section{AgentEval Task Criteria} \label{sec:agenteval_criteria}

\begin{table*}[tbh]
\centering
\resizebox{\linewidth}{!}{
\begin{tabular}{lll}
\hline
\multicolumn{1}{l|}{\textbf{Criteria}} & \multicolumn{1}{l|}{\textbf{Description}} & \textbf{Accepted Values} \\ \hline

\multicolumn{3}{l}{\textbf{Task: Relevance Judgment}} \\ \hline

\multicolumn{1}{l|}{\textbf{Clarity}} & 
\multicolumn{1}{l|}{\begin{tabular}[c]{@{}l@{}}The degree to which the reasoning behind the relevance \\ judgment is clear and understandable.\end{tabular}} & 
\begin{tabular}[c]{@{}l@{}}1 - very unclear, 2 - unclear, \\ 3 - neutral, 4 - clear, \\ 5 - very clear\end{tabular} \\ \hline

\multicolumn{1}{l|}{\textbf{Contextuality}} & 
\multicolumn{1}{l|}{\begin{tabular}[c]{@{}l@{}}The extent to which the judgment considers the specific context \\ of the conversation including relationships, settings, and \\ cultural nuances.\end{tabular}} & 
\begin{tabular}[c]{@{}l@{}}1 - not contextualized, 2 - poorly contextualized, \\ 3 - moderately contextualized, 4 - well contextualized, \\ 5 - very well contextualized\end{tabular} \\ \hline

\multicolumn{1}{l|}{\textbf{Appropriateness}} & 
\multicolumn{1}{l|}{\begin{tabular}[c]{@{}l@{}}How well the social norm applies to the situation \\ discussed in the conversation.\end{tabular}} & 
\begin{tabular}[c]{@{}l@{}}1 - not appropriate, 2 - minimally appropriate, \\ 3 - somewhat appropriate, 4 - appropriate, \\ 5 - highly appropriate\end{tabular} \\ \hline

\multicolumn{1}{l|}{\textbf{Cultural Relevance}} & 
\multicolumn{1}{l|}{\begin{tabular}[c]{@{}l@{}}The degree to which the social norm reflects significant \\ aspects of Chinese culture relevant to the conversation.\end{tabular}} & 
\begin{tabular}[c]{@{}l@{}}1 - not culturally relevant, 2 - minimally relevant, \\ 3 - somewhat relevant, 4 - culturally relevant, \\ 5 - highly culturally relevant\end{tabular} \\ \hline

\multicolumn{1}{l|}{\textbf{Consistency}} & 
\multicolumn{1}{l|}{\begin{tabular}[c]{@{}l@{}}The consistency of the judgment with established norms \\ and expectations in Chinese cultural contexts.\end{tabular}} & 
\begin{tabular}[c]{@{}l@{}}1 - highly inconsistent, 2 - inconsistent, \\ 3 - neutral, 4 - consistent, \\ 5 - highly consistent\end{tabular} \\ \hline

\multicolumn{1}{l|}{\textbf{Relationship Dynamics}} & 
\multicolumn{1}{l|}{\begin{tabular}[c]{@{}l@{}}The consideration of the relationships between the individuals \\ involved in the conversation and how that influences \\ the relevance of the social norm.\end{tabular}} & 
\begin{tabular}[c]{@{}l@{}}1 - not considered, 2 - weakly considered, \\ 3 - moderately considered, 4 - strongly considered, \\ 5 - very strongly considered\end{tabular} \\ \hline

\multicolumn{1}{l|}{\textbf{Setting Analysis}} & 
\multicolumn{1}{l|}{\begin{tabular}[c]{@{}l@{}}Evaluation of how the setting of the conversation (e.g., family, \\ workplace, casual gathering) impacts the relevance \\ of the social norm.\end{tabular}} & 
\begin{tabular}[c]{@{}l@{}}1 - no impact, 2 - minor impact, \\ 3 - moderate impact, 4 - significant impact, \\ 5 - critical impact\end{tabular} \\ \hline

\multicolumn{1}{l|}{\textbf{Norm Visibility}} & 
\multicolumn{1}{l|}{\begin{tabular}[c]{@{}l@{}}The extent to which the social norm is visible or recognized \\ in the context of the conversation.\end{tabular}} & 
\begin{tabular}[c]{@{}l@{}}1 - not visible, 2 - slightly visible, \\ 3 - moderately visible, 4 - visible, \\ 5 - very visible\end{tabular} \\ \hline

\multicolumn{1}{l|}{\textbf{Social Hierarchy}} & 
\multicolumn{1}{l|}{\begin{tabular}[c]{@{}l@{}}How well the judgment incorporates aspects of social hierarchy \\ or status within the Chinese cultural context.\end{tabular}} & 
\begin{tabular}[c]{@{}l@{}}1 - not addressed, 2 - poorly addressed, \\ 3 - addressed, 4 - well addressed, \\ 5 - very well addressed\end{tabular} \\ \hline

\multicolumn{1}{l|}{\textbf{Timeliness}} & 
\multicolumn{1}{l|}{\begin{tabular}[c]{@{}l@{}}Consider how the relevance of the social norm may change \\ over time or in different historical contexts.\end{tabular}} & 
\begin{tabular}[c]{@{}l@{}}1 - not timely, 2 - slightly timely, \\ 3 - moderately timely, 4 - timely, \\ 5 - very timely\end{tabular} \\ \hline

\multicolumn{1}{l|}{\textbf{Evidence}} & 
\multicolumn{1}{l|}{\begin{tabular}[c]{@{}l@{}}The amount and quality of evidence or examples provided \\ to support the relevance judgment.\end{tabular}} & 
\begin{tabular}[c]{@{}l@{}}1 - no evidence, 2 - minimal evidence, \\ 3 - some evidence, 4 - good evidence, \\ 5 - strong evidence\end{tabular} \\ \hline

\multicolumn{1}{l|}{\textbf{Norm Specificity}} & 
\multicolumn{1}{l|}{\begin{tabular}[c]{@{}l@{}}The degree to which the social norm discussed is specific \\ and detailed rather than vague.\end{tabular}} & 
\begin{tabular}[c]{@{}l@{}}1 - very vague, 2 - vague, \\ 3 - somewhat specific, 4 - specific, \\ 5 - very specific\end{tabular} \\ \hline

\multicolumn{1}{l|}{\textbf{Emotional Tone}} & 
\multicolumn{1}{l|}{\begin{tabular}[c]{@{}l@{}}Evaluation of how the emotional tone of the conversation \\ affects the relevance of the social norm.\end{tabular}} & 
\begin{tabular}[c]{@{}l@{}}1 - negative impact, 2 - slight negative impact, \\ 3 - neutral impact, 4 - slight positive impact, \\ 5 - positive impact\end{tabular} \\ \hline

\multicolumn{1}{l|}{\textbf{Discourse Style}} & 
\multicolumn{1}{l|}{\begin{tabular}[c]{@{}l@{}}The impact of the discourse style (formal, informal, persuasive, etc.) \\ on the relevance of the social norm.\end{tabular}} & 
\begin{tabular}[c]{@{}l@{}}1 - no impact, 2 - minimal impact, \\ 3 - moderate impact, 4 - significant impact, \\ 5 - critical impact\end{tabular} \\ \hline

\multicolumn{1}{l|}{\textbf{Dissonance Level}} & 
\multicolumn{1}{l|}{\begin{tabular}[c]{@{}l@{}}The level of dissonance between the social norm and the expressed \\ opinions or behaviors in the conversation.\end{tabular}} & 
\begin{tabular}[c]{@{}l@{}}1 - highly dissonant, 2 - moderately dissonant, \\ 3 - neutral, 4 - somewhat aligned, \\ 5 - highly aligned\end{tabular} \\ \hline

\multicolumn{1}{l|}{\textbf{Cultural Evolution}} & 
\multicolumn{1}{l|}{\begin{tabular}[c]{@{}l@{}}Consideration of how modern trends or changes in society may \\ affect the relevance of traditional social norms.\end{tabular}} & 
\begin{tabular}[c]{@{}l@{}}1 - outdated, 2 - slightly outdated, \\ 3 - somewhat relevant, 4 - relevant, \\ 5 - highly relevant\end{tabular} \\ \hline

\multicolumn{1}{l|}{\textbf{Collective Perspective}} & 
\multicolumn{1}{l|}{\begin{tabular}[c]{@{}l@{}}How well the judgment considers the views of the collective society \\ versus individual perspectives.\end{tabular}} & 
\begin{tabular}[c]{@{}l@{}}1 - individual-focused, 2 - slightly collective, \\ 3 - moderately collective, 4 - largely collective, \\ 5 - entirely collective\end{tabular} \\ \hline

\multicolumn{1}{l|}{\textbf{Norm Adoption}} & 
\multicolumn{1}{l|}{\begin{tabular}[c]{@{}l@{}}Recognition of how widely the social norm is adopted or practiced \\ within Chinese society.\end{tabular}} & 
\begin{tabular}[c]{@{}l@{}}1 - not adopted, 2 - rarely adopted, \\ 3 - somewhat adopted, 4 - widely adopted, \\ 5 - universally adopted\end{tabular} \\ \hline

\multicolumn{1}{l|}{\textbf{Age Sensitivity}} & 
\multicolumn{1}{l|}{\begin{tabular}[c]{@{}l@{}}The extent to which the relevance judgment is sensitive to the \\ age differences of the individuals involved.\end{tabular}} & 
\begin{tabular}[c]{@{}l@{}}1 - not sensitive, 2 - slightly sensitive, \\ 3 - moderately sensitive, 4 - sensitive, \\ 5 - very sensitive\end{tabular} \\ \hline

\multicolumn{1}{l|}{\textbf{Analytic Depth}} & 
\multicolumn{1}{l|}{\begin{tabular}[c]{@{}l@{}}The depth of analysis provided in the reasoning for the \\ relevance judgment.\end{tabular}} & 
\begin{tabular}[c]{@{}l@{}}1 - very superficial, 2 - superficial, \\ 3 - moderate depth, 4 - deep, \\ 5 - very deep\end{tabular} \\ \hline

\multicolumn{1}{l|}{\textbf{Feedback Responsiveness}} & 
\multicolumn{1}{l|}{\begin{tabular}[c]{@{}l@{}}The degree to which the judgment accounts for feedback or \\ reactions from the individuals in the conversation.\end{tabular}} & 
\begin{tabular}[c]{@{}l@{}}1 - not responsive, 2 - slightly responsive, \\ 3 - moderately responsive, 4 - responsive, \\ 5 - very responsive\end{tabular} \\ \hline

\multicolumn{1}{l|}{\textbf{Rule Conformity}} & 
\multicolumn{1}{l|}{\begin{tabular}[c]{@{}l@{}}How closely the judgment conforms to established social rules \\ and expectations within the context.\end{tabular}} & 
\begin{tabular}[c]{@{}l@{}}1 - highly non-conformant, 2 - non-conformant, \\ 3 - somewhat conformant, 4 - conformant, \\ 5 - highly conformant\end{tabular} \\ \hline

\end{tabular}
}
\caption{Evaluation Criteria Generated by AgentEval for \textit{Relevance} of \textit{Norm Description} to \textit{Conversation} Judgment}
\label{tab:task_criteria_relevance}
\end{table*}

\begin{table*}[tbh]
\centering
\resizebox{0.8\linewidth}{!}{
\begin{tabular}{lll}
\hline
\multicolumn{1}{l|}{\textbf{Criteria}} & \multicolumn{1}{l|}{\textbf{Description}} & \textbf{Accepted Values} \\ \hline

\multicolumn{3}{l}{\textbf{Task: Norm Interpretation and Evaluation}} \\ \hline

\multicolumn{1}{l|}{\textbf{Contextual Influence Evaluation}} & 
\multicolumn{1}{l|}{\begin{tabular}[c]{@{}l@{}}Evaluate how the context (e.g., setting, relationships) \\ influences the interpretation of the norm.\end{tabular}} & 
\begin{tabular}[c]{@{}l@{}}high influence, \\ medium influence, \\ low influence, \\ no influence\end{tabular} \\ \hline

\multicolumn{1}{l|}{\textbf{Cultural Appropriateness Assessment}} & 
\multicolumn{1}{l|}{\begin{tabular}[c]{@{}l@{}}Judge whether the norm is appropriate within \\ the cultural context of the conversation.\end{tabular}} & 
\begin{tabular}[c]{@{}l@{}}appropriate, \\ inappropriate\end{tabular} \\ \hline

\multicolumn{1}{l|}{\textbf{Generational Perspective Evaluation}} & 
\multicolumn{1}{l|}{\begin{tabular}[c]{@{}l@{}}Assess how generational differences impact \\ the conversation's norms.\end{tabular}} & 
\begin{tabular}[c]{@{}l@{}}significant difference, \\ some difference, \\ no difference\end{tabular} \\ \hline

\multicolumn{1}{l|}{\textbf{Emotional Weight Assessment}} & 
\multicolumn{1}{l|}{\begin{tabular}[c]{@{}l@{}}Determine the emotional significance of \\ the norm within the conversation.\end{tabular}} & 
\begin{tabular}[c]{@{}l@{}}high weight, \\ medium weight, \\ low weight\end{tabular} \\ \hline

\multicolumn{1}{l|}{\textbf{Implications of Norm Violation}} & 
\multicolumn{1}{l|}{\begin{tabular}[c]{@{}l@{}}Assess the potential consequences of violating \\ the social norm.\end{tabular}} & 
\begin{tabular}[c]{@{}l@{}}severe consequences, \\ moderate consequences, \\ mild consequences, \\ no consequences\end{tabular} \\ \hline

\multicolumn{1}{l|}{\textbf{Responsibility Assessment}} & 
\multicolumn{1}{l|}{\begin{tabular}[c]{@{}l@{}}Evaluate who holds the primary responsibility \\ for adhering to the social norm in the conversation.\end{tabular}} & 
\begin{tabular}[c]{@{}l@{}}enactor, \\ acceptor, \\ both\end{tabular} \\ \hline

\multicolumn{1}{l|}{\textbf{Social Norm Awareness Evaluation}} & 
\multicolumn{1}{l|}{\begin{tabular}[c]{@{}l@{}}Judge the awareness of the social norm \\ by the involved parties.\end{tabular}} & 
\begin{tabular}[c]{@{}l@{}}fully aware, \\ somewhat aware, \\ not aware\end{tabular} \\ \hline

\multicolumn{1}{l|}{\textbf{Feedback Necessity Assessment}} & 
\multicolumn{1}{l|}{\begin{tabular}[c]{@{}l@{}}Determine if feedback is necessary based on \\ the adherence or violation of the norm.\end{tabular}} & 
\begin{tabular}[c]{@{}l@{}}necessary, \\ not necessary\end{tabular} \\ \hline

\multicolumn{1}{l|}{\textbf{Coping Strategy Evaluation}} & 
\multicolumn{1}{l|}{\begin{tabular}[c]{@{}l@{}}Evaluate the coping strategies employed by individuals \\ in response to a norm violation.\end{tabular}} & 
\begin{tabular}[c]{@{}l@{}}effective, \\ somewhat effective, \\ ineffective\end{tabular} \\ \hline

\multicolumn{1}{l|}{\textbf{Communication Style Analysis}} & 
\multicolumn{1}{l|}{\begin{tabular}[c]{@{}l@{}}Analyze the communication styles used \\ in relation to the social norm.\end{tabular}} & 
\begin{tabular}[c]{@{}l@{}}formal, \\ informal, \\ assertive, \\ passive, \\ aggressive\end{tabular} \\ \hline

\end{tabular}
}
\caption{Evaluation Criteria Generated by AgentEval for \textit{Symbolic Annotation Quality} Judgment}
\label{tab:task_criteria_symbolic_annotation}
\end{table*}



\section{Data Sources and Statistics} \label{sec:pipeline_data_statistics}
\begin{table*}[!tbhp]
\resizebox{\textwidth}{!}{
\begin{tabular}{|l|cccc|cccccccccc|cc|}
\hline
\multirow{3}{*}{\textbf{Dataset}} & \multicolumn{4}{c|}{\textbf{Turn-Level}}                                                                                                                                                                                                                  & \multicolumn{10}{c|}{\textbf{Conversation-Level}}                                                                                                                                                                                                                                                                                                                                                                                                                                                                                                                                                                                                                                                                                                                                                                                 & \multicolumn{2}{c|}{\textbf{Corpus}}                                                                                                                                              \\ \cline{2-17} 
                                  & \multicolumn{1}{c|}{\multirow{2}{*}{\textbf{Em.}}} & \multicolumn{1}{c|}{\multirow{2}{*}{\textbf{DA}}} & \multicolumn{1}{c|}{\multirow{2}{*}{\textbf{Sent.}}} & \multirow{2}{*}{\textbf{\begin{tabular}[c]{@{}c@{}}Sp.-List.\\ Reln.\end{tabular}}} & \multicolumn{1}{c|}{\multirow{2}{*}{\textbf{Summ.}}} & \multicolumn{3}{c|}{\textbf{Norms}}                                                                                                                                                                                                                             & \multicolumn{3}{c|}{\textbf{Violations}}                                                                                                                                                                                                                      & \multicolumn{2}{c|}{\textbf{Effects}}                                                                                                                                & \multirow{2}{*}{\textbf{Field}}                         & \multicolumn{2}{c|}{\textbf{\begin{tabular}[c]{@{}c@{}}Norm\\ Concepts\end{tabular}}}                                                                                             \\ \cline{7-14} \cline{16-17} 
                                  & \multicolumn{1}{c|}{}                              & \multicolumn{1}{c|}{}                             & \multicolumn{1}{c|}{}                                &                                                                                     & \multicolumn{1}{c|}{}                                & \multicolumn{1}{c|}{\textbf{Desc}}                                          & \multicolumn{1}{c|}{\textbf{Valid}}                                                     & \multicolumn{1}{c|}{\textbf{\begin{tabular}[c]{@{}c@{}}Symb.\\ Attr\end{tabular}}}      & \multicolumn{1}{c|}{\textbf{Desc}}                                          & \multicolumn{1}{c|}{\textbf{Valid}}                                                    & \multicolumn{1}{c|}{\textbf{\begin{tabular}[c]{@{}c@{}}Symb.\\ Attr\end{tabular}}}     & \multicolumn{1}{c|}{\textbf{Desc}}                                          & \multicolumn{1}{c|}{\textbf{Valid}}                                                    &                                                         & \multicolumn{1}{c|}{\textbf{Conc.}}                                                    & \textbf{\begin{tabular}[c]{@{}c@{}}Assg.\\ Norms\end{tabular}}                           \\ \hline
\textbf{MPDD}                     & \multicolumn{1}{c|}{A}                             & \multicolumn{1}{c|}{LM}                           & \multicolumn{1}{c|}{-}                               & A                                                                                   & \multicolumn{1}{c|}{GPT}                             & \multicolumn{1}{c|}{\begin{tabular}[c]{@{}c@{}}10,637\\ (GPT)\end{tabular}} & \multicolumn{1}{c|}{\multirow{3}{*}{\begin{tabular}[c]{@{}c@{}}900\\ (H)\end{tabular}}} & \multicolumn{1}{c|}{\multirow{3}{*}{\begin{tabular}[c]{@{}c@{}}726\\ (H)\end{tabular}}} & \multicolumn{1}{c|}{\begin{tabular}[c]{@{}c@{}}5,721\\ (GPT)\end{tabular}}  & \multicolumn{1}{c|}{\multirow{3}{*}{\begin{tabular}[c]{@{}c@{}}213+65915\\ (H+GPT)\end{tabular}}} & \multicolumn{1}{c|}{\multirow{3}{*}{\begin{tabular}[c]{@{}c@{}}\\ 66,128\\(H+GPT)\end{tabular}}} & \multicolumn{1}{c|}{\begin{tabular}[c]{@{}c@{}}14,521\\ (GPT)\end{tabular}} & \multicolumn{1}{c|}{\multirow{3}{*}{\begin{tabular}[c]{@{}c@{}}213\\ (H)\end{tabular}}} & \begin{tabular}[c]{@{}c@{}}4,141\\ (A)\end{tabular}     & \multicolumn{1}{c|}{\multirow{3}{*}{\begin{tabular}[c]{@{}c@{}}35\\ (H)\end{tabular}}} & \multirow{3}{*}{\begin{tabular}[c]{@{}c@{}}422\\ (H)\\ \\ 36,954 \\ (k-NN)\end{tabular}} \\ \cline{1-7} \cline{10-10} \cline{13-13} \cline{15-15}
\textbf{CPED}                     & \multicolumn{1}{c|}{A}                             & \multicolumn{1}{c|}{A}                            & \multicolumn{1}{c|}{A}                               & LM                                                                                  & \multicolumn{1}{c|}{GPT}                             & \multicolumn{1}{c|}{\begin{tabular}[c]{@{}c@{}}32,209\\ (GPT)\end{tabular}} & \multicolumn{1}{c|}{}                                                                   & \multicolumn{1}{c|}{}                                                                   & \multicolumn{1}{c|}{\begin{tabular}[c]{@{}c@{}}17,865\\ (GPT)\end{tabular}} & \multicolumn{1}{c|}{}                                                                  & \multicolumn{1}{c|}{}                                                                  & \multicolumn{1}{c|}{\begin{tabular}[c]{@{}c@{}}25,866\\ (GPT)\end{tabular}} & \multicolumn{1}{c|}{}                                                                  & \begin{tabular}[c]{@{}c@{}}11,832\\ (A)\end{tabular}    & \multicolumn{1}{c|}{}                                                                  &                                                                                          \\ \cline{1-7} \cline{10-10} \cline{13-13} \cline{15-15}
\textbf{LDC CCU}                  & \multicolumn{1}{c|}{A}                             & \multicolumn{1}{c|}{A}                            & \multicolumn{1}{c|}{-}                               & LM                                                                                  & \multicolumn{1}{c|}{GPT}                             & \multicolumn{1}{c|}{\begin{tabular}[c]{@{}c@{}}23,306\\ (GPT)\end{tabular}} & \multicolumn{1}{c|}{}                                                                   & \multicolumn{1}{c|}{}                                                                   & \multicolumn{1}{c|}{\begin{tabular}[c]{@{}c@{}}15,605\\ (GPT)\end{tabular}} & \multicolumn{1}{c|}{}                                                                  & \multicolumn{1}{c|}{}                                                                  & \multicolumn{1}{c|}{\begin{tabular}[c]{@{}c@{}}14,908\\ (GPT)\end{tabular}} & \multicolumn{1}{c|}{}                                                                  & \begin{tabular}[c]{@{}c@{}}7,554\\ (A, LM)\end{tabular} & \multicolumn{1}{c|}{}                                                                  &                                                                                          \\ \hline
\end{tabular}
}
\caption{Sources and Counts of Collected Schema Grounding Dataset. A - gold annotation; LM - Llama-70b generated; GPT - GPT-3.5 generated; H - human annotation; kNN - interactive k-nearest neighbors; Em - emotion; DA - dialogue act; Sent - sentiment; Sp\_List Reln - speaker\_listener relationship; Desc - descriptions; Symb\_Attr - symbolic attributes; Valid - Validated; Conc - concepts; Assg\_Norms - norms assigned to concepts;\label{tab:data_sources}}
\end{table*}

\section{Annotation GUI}\label{sec:annotation_gui}

\begin{figure*}[htbp]
    \centering
    \includegraphics[width=\textwidth]{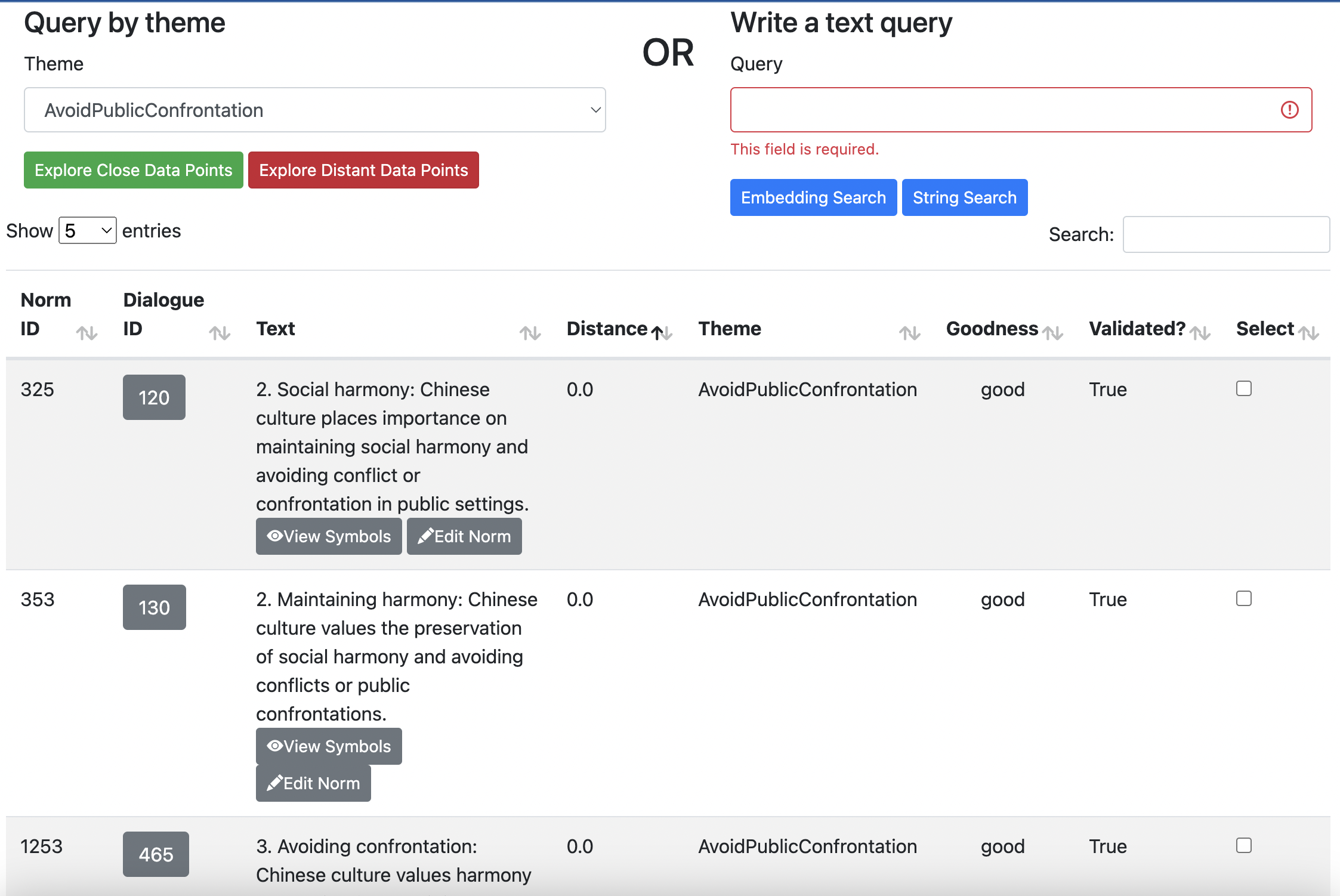}
    \caption{Annotation Interface for Norm Concept Discovery}
    \label{fig:ann_ss1}
\end{figure*} 

\begin{figure*}[htbp]
    \centering
    \includegraphics[width=\textwidth]{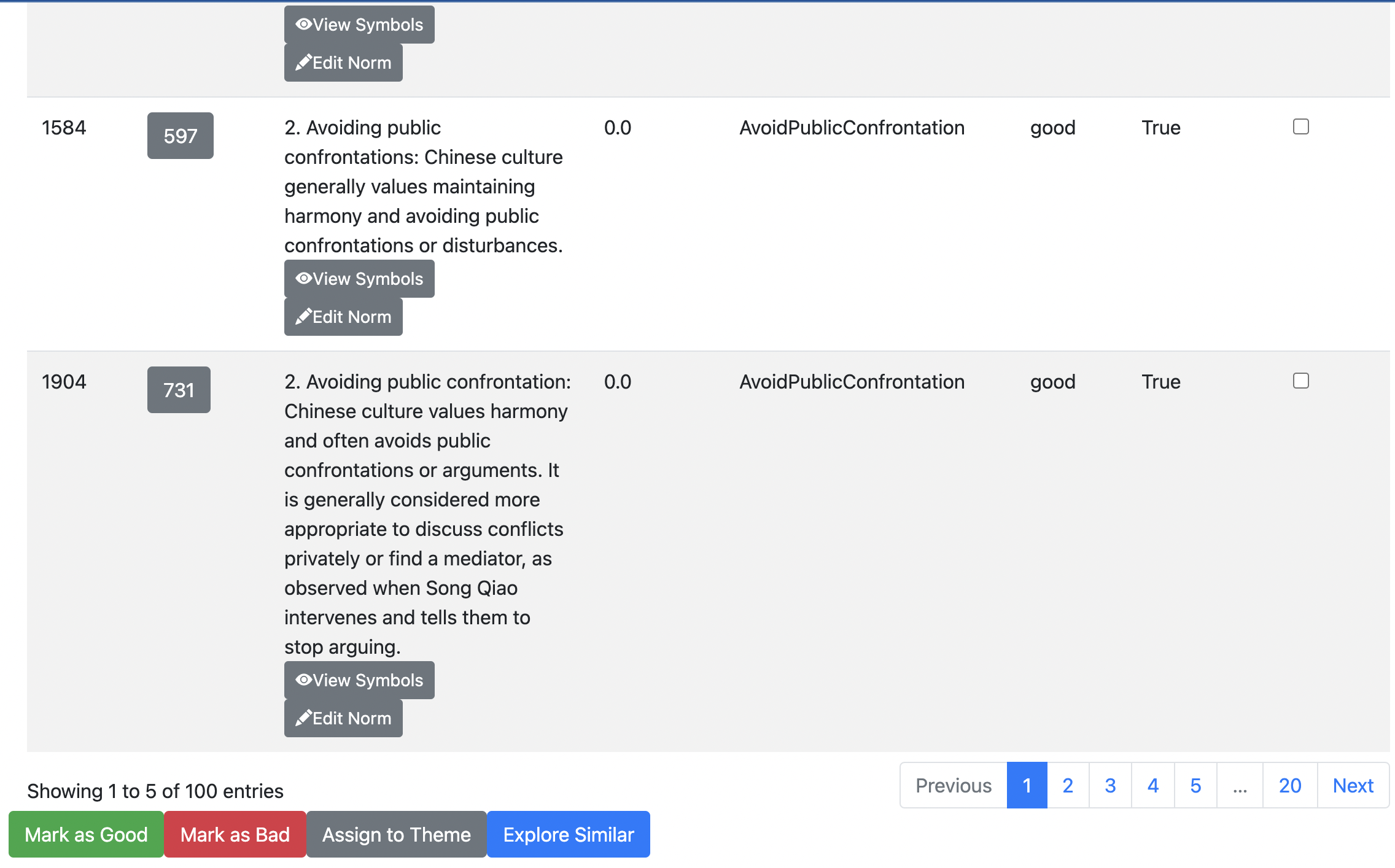}
    \caption{Annotation Interface for Norm Concept Discovery}
    \label{fig:ann_ss2}
\end{figure*}

\begin{figure*}[htbp]
    \centering
    \includegraphics[width=\textwidth]{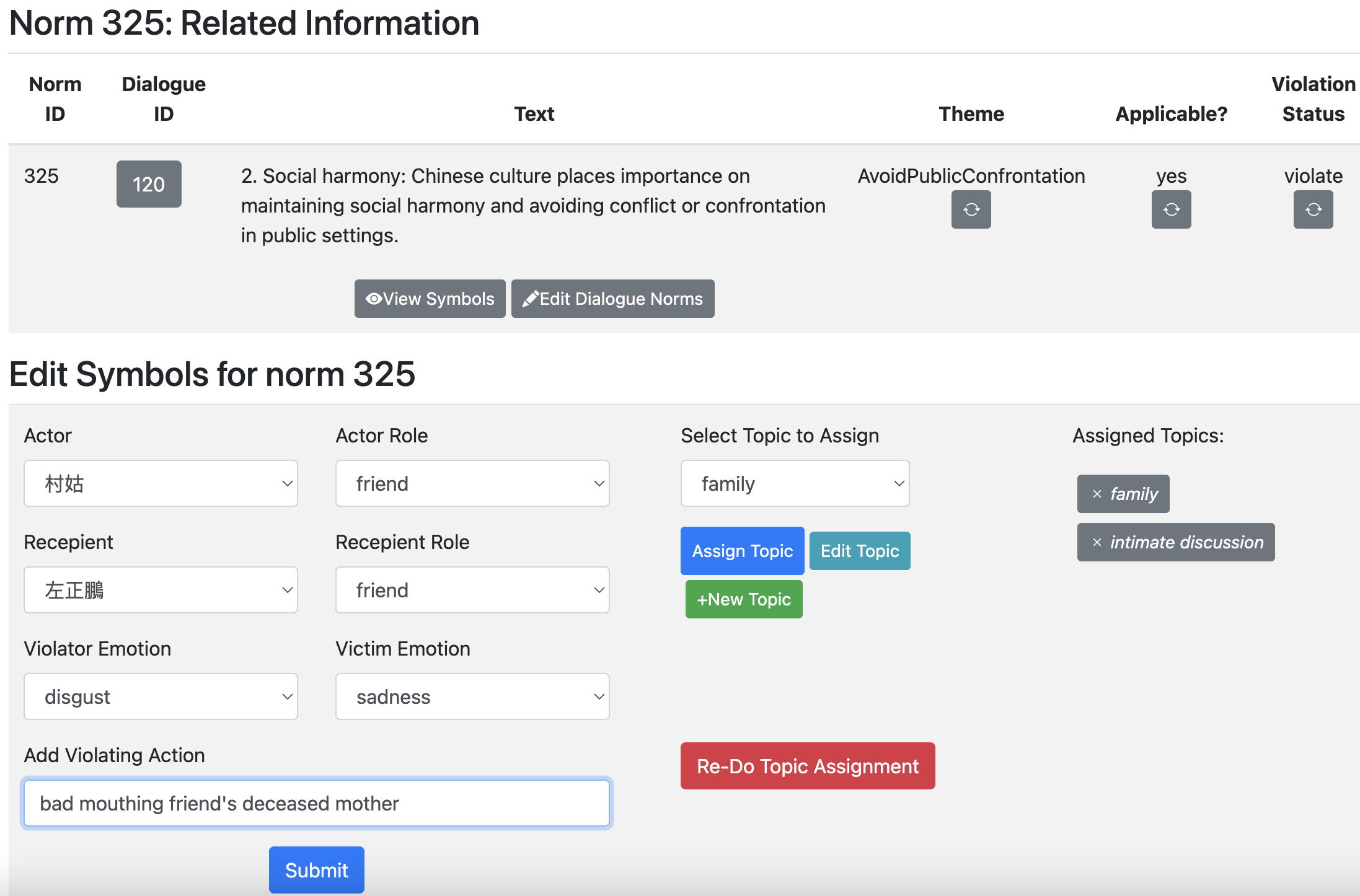}
    \caption{Human Validation and Symbolic Annotation of Cultural Context Annotation Framework}
    \label{fig:human_validation}
\end{figure*}

\section{Norm Concept Visualization}

\begin{figure*}[tbh]
    \includegraphics[width=\textwidth]{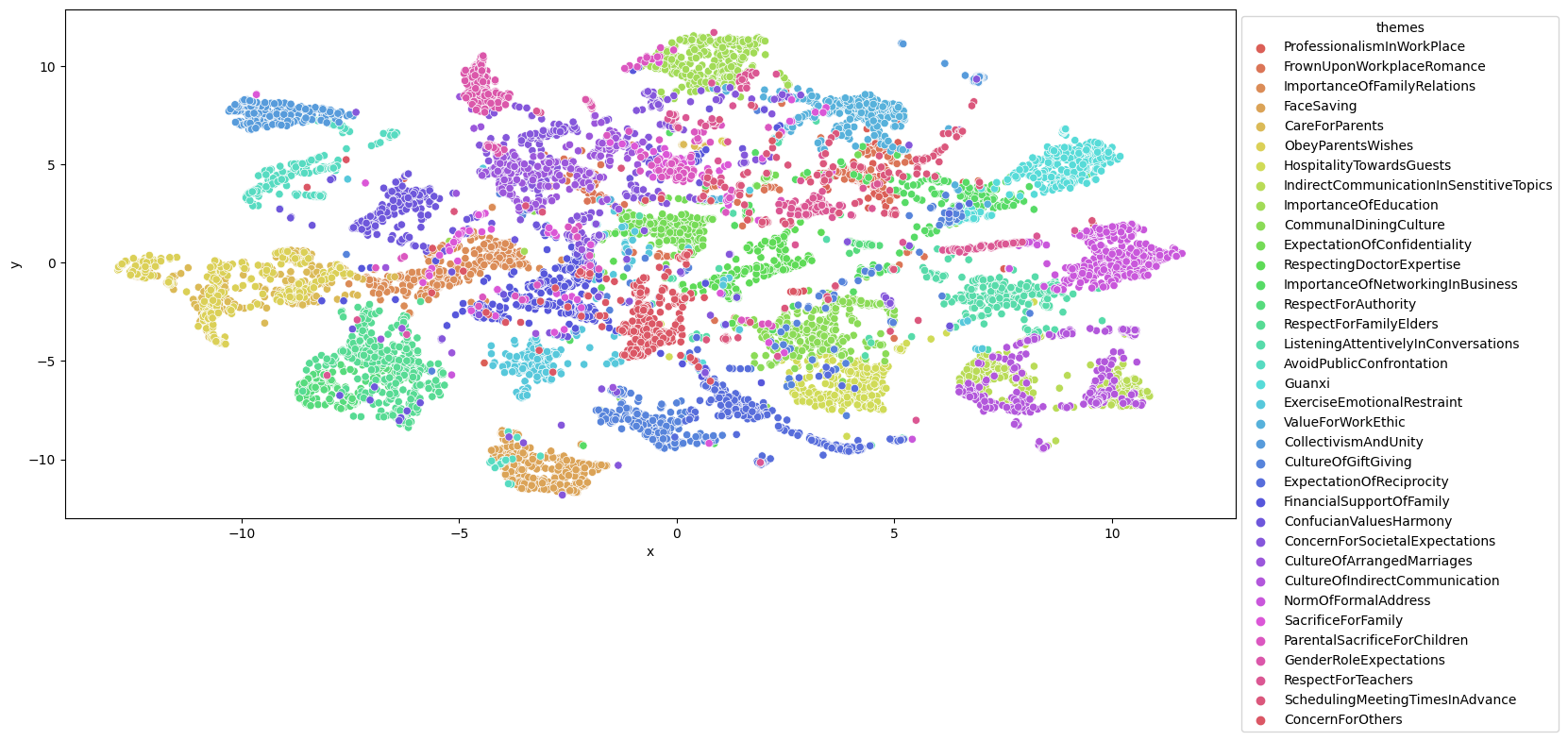}
    \caption{Norm Concept Visualization}
\end{figure*}

\section{Schema Example}
\begin{figure*}
    \centering
    \includegraphics[width=0.75\textwidth]{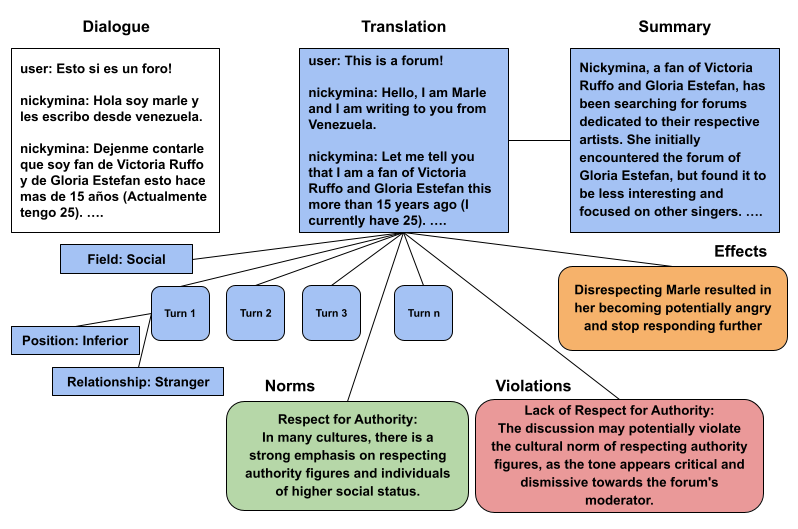}
    \label{fig:schema_example}
    \caption{An Example Instance of Schema Augmented Conversation}
\end{figure*}

\end{document}